\PassOptionsToPackage{table}{xcolor}
\documentclass[sigconf,authorversion]{acmart}
\usepackage{listings}

\usepackage{CJKutf8}
\usepackage{booktabs}
\usepackage{multicol,multirow}
\usepackage{algorithm2e}
\usepackage{subcaption} 

\AtBeginDocument{%
  \providecommand\BibTeX{{%
    \normalfont B\kern-0.5em{\scshape i\kern-0.25em b}\kern-0.8em\TeX}}}

\setcopyright{acmcopyright}

\copyrightyear{2023}
\acmYear{2023}
\setcopyright{acmlicensed}\acmConference[CIKM '23]{Proceedings of the 32nd ACM International Conference on Information and Knowledge Management}{October 21--25, 2023}{Birmingham, United Kingdom}
\acmBooktitle{Proceedings of the 32nd ACM International Conference on Information and Knowledge Management (CIKM '23), October 21--25, 2023, Birmingham, United Kingdom}
\acmPrice{15.00}
\acmDOI{10.1145/3583780.3614867}
\acmISBN{979-8-4007-0124-5/23/10}

\begin{document}
\begin{CJK}{UTF8}{mj}

\newcommand{\todo}[1]{\textcolor{red}{#1}}
\newcommand{\tore}[1]{\textcolor{orange}{#1}}
\newcommand{\tohi}[1]{\textcolor{blue}{#1}}

\newcommand{\js}[1]{\textsf{\textcolor{magenta}{#1}}}
\newcommand{\cm}[1]{\textcolor{red}{#1}}
\newcommand{\bm}[1]{\textcolor{blue}{#1}}



\title{Enhancing Spatiotemporal Traffic Prediction through Urban Human Activity Analysis}




\author{Sumin Han}
\orcid{0000-0002-4071-8469}
\affiliation{%
  \department{School of Computing}
  \institution{Korea Advanced Institute of Science and Technology}
  \streetaddress{291 Daehak-ro, Yuseong-gu}
  \city{Daejeon}
  \country{Republic of Korea}
  \postcode{34141}
}
\email{hsm6911@kaist.ac.kr}

\author{Youngjun Park}
\orcid{0000-0002-4254-2268}
\affiliation{%
  \department{School of Computing}
  \institution{Korea Advanced Institute of Science and Technology}
  \streetaddress{291 Daehak-ro, Yuseong-gu}
  \city{Daejeon}
  \country{Republic of Korea}
  \postcode{34141}
}
\email{youngjourpark@kaist.ac.kr}

\author{Minji Lee}
\orcid{0000-0001-5293-8904}
\affiliation{%
  \department{School of Computing}
  \institution{Korea Advanced Institute of Science and Technology}
  \streetaddress{291 Daehak-ro, Yuseong-gu}
  \city{Daejeon}
  \country{Republic of Korea}
  \postcode{34141}
}
\email{haewon\_lee@kaist.ac.kr}

\author{Jisun An}
\orcid{0000-0002-4353-8009}
\affiliation{%
  \department{Luddy School of Informatics, Computing, and, Engineering}
  \institution{Indiana University Bloomington}
  \city{Bloomington}
  \country{USA}
}
\email{jisunan@iu.edu}

\author{Dongman Lee}
\orcid{0000-0001-5923-6227}
\affiliation{%
  \department{School of Computing}
  \institution{Korea Advanced Institute of Science and Technology}
  \streetaddress{291 Daehak-ro, Yuseong-gu}
  \city{Daejeon}
  \country{Republic of Korea}
  \postcode{34141}
}
\email{dlee@kaist.ac.kr}



\begin{abstract}

Traffic prediction is one of the key elements to ensure the safety and convenience of citizens. Existing traffic prediction models primarily focus on deep learning architectures to capture spatial and temporal correlation. They often overlook the underlying nature of traffic. Specifically, the sensor networks in most traffic datasets do not accurately represent the actual road network exploited by vehicles, failing to provide insights into the traffic patterns in urban activities. To overcome these limitations, we propose an improved traffic prediction method based on graph convolution deep learning algorithms. We leverage human activity frequency data from National Household Travel Survey to enhance the inference capability of a causal relationship between activity and traffic patterns. Despite making minimal modifications to the conventional graph convolutional recurrent networks and graph convolutional transformer architectures, our approach achieves state-of-the-art performance without introducing excessive computational overhead.

\end{abstract}



\maketitle

\section{Introduction}

Traffic prediction plays a crucial role in ensuring the safety and convenience of citizens by accurately estimating future traffic speed or volume based on historical patterns from sensor data. 
Unlike typical time-series prediction problems, traffic prediction requires inferring a sensor's traffic values by leveraging patterns observed in other sensors. 
Previous studies have explored various spatiotemporal models based on Recurrent Neural Networks (RNN) \cite{DCRNN, STGCN, ASTGCN, T-GCN} and Transformer \cite{ASTGCNNTF,STEP,PDFormer}, which have shown effectiveness in time series prediction while incorporating the spatial similarity between traffic sensors.

\begin{figure}[t]
  \includegraphics[width=\columnwidth]{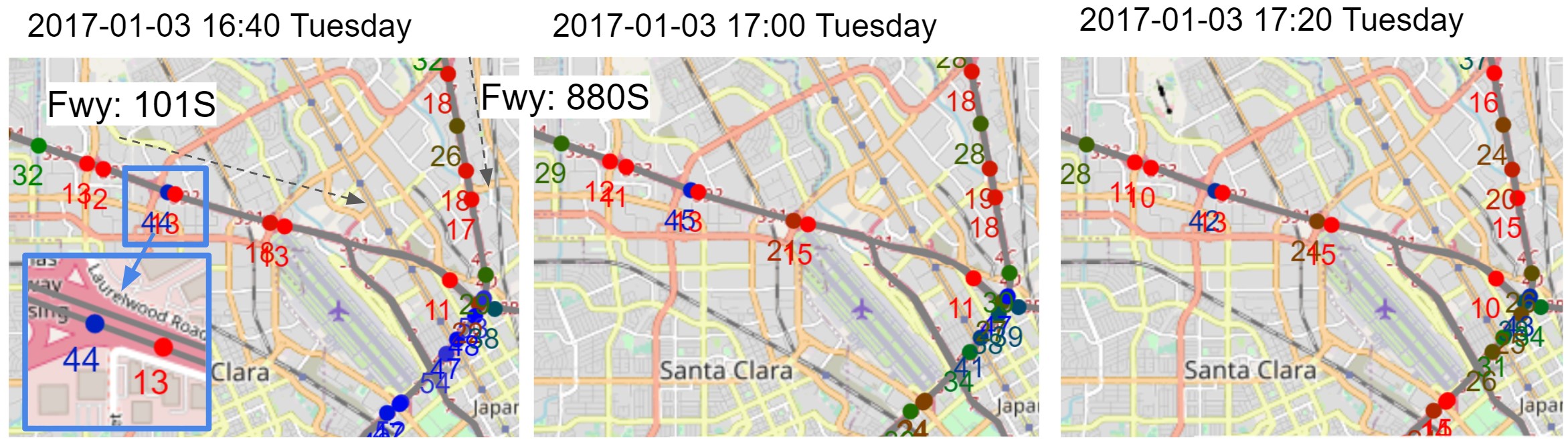}
  \vspace{-5mm}
  \caption{Congestion pattern of sensors on specific freeways in PEMS-BAY dataset (congestion level: red>green>blue).}
  \label{fig:showcase}
  \vspace{-3mm}
\end{figure}

We argue that there still remain key points for improving traffic prediction:

\textbf{1. Construction of an accurate graph representation of the sensor network:}
Existing studies \cite{DCRNN,STGCN,T-GCN} construct adjacency matrices based on sensor proximity using distance thresholds. 
Although some models \cite{GTS,GraphWavenet,ASTGCNNTF,STEP} have attempted to optimize the sensor adjacency matrix by learning from data, these approaches often result in inefficiency and high computational costs. Additionally, they may generate adjacency matrices that include artificial connections, leading to inaccuracies and incorrect representations of the sensor network.
For example, in situations where distant traffic sensors show similar traffic patterns, machine learning models might mistakenly deduce a meaningful connection between them, despite the absence of any causal relationship.

\textbf{2. Addressing individual sensor spatial heterogeneity:} \label{sec:need-sensor-embedding}
Each traffic sensor is situated within a unique built environment, resulting in diverse congestion patterns. 
For instance, congestion due to rush hour may occur only in specific lanes or sensors.
Even in close locations, different patterns can emerge due to factors such as the number of sensor lanes, entry and exit lanes, and installation positions. Fig.~\ref{fig:showcase} illustrates a significant discrepancy in values between adjacent sensors, where a sensor recording a speed of 44 miles per hour (mph) could be considered congested relative to neighboring sensors. While previous work such as \cite{ASTGCNNTF} has addressed this issue by leveraging spatial positional encoding for Transformer models, the primary focus has been on handling positional encoding rather than normalizing the patterns of individual sensors.

\textbf{3. Incorporating human activity-based inference for traffic prediction:}
Human activities, such as commuting, significantly influence traffic patterns and can lead to congestion. Previous studies have incorporated temporal information such as weekday and time-of-day to capture correlations between time and traffic patterns~\cite{ASTGCNNTF,GMAN,PDFormer}.
While temporal information provides insights into human activities, it does not establish direct causality, as traffic patterns are influenced by human actions.
Indirectly inferring human actions from temporal information makes it challenging for models to detect variations in behavior, such as during holidays or seasonal activity difference, or learn about similar behaviors occurring at different times.

In this study, we propose a novel solution to address the challenges of generating realistic vehicle travel trajectories while ensuring sensor connectivity. We utilize the A* algorithm to generate these trajectories and derive pairwise similarity measures between sensors, integrating them into graph convolutional temporal models. 
To accommodate the spatial heterogeneity of sensors, we employ a one-hot-based sensor encoding specific to each sensor, enabling adaptability to diverse sensor environments. 
To capture the correlation between human activity and traffic patterns, we incorporate urban travel activity frequencies from the National Household Travel Survey\cite{nths-2017} estimated at the target prediction timestamp. Our approach consists of two spatiotemporal deep learning architectures, namely UAGCRN and UAGCTransformer, which effectively integrate the constructed graph on graph-convolutional recurrent neural networks and graph-convolutional transformers. Our model surpasses other baselines and achieves state-of-the-art performance on conventional traffic datasets. We demonstrate the superiority of our constructed graph by comparing its impact on other spatiotemporal models. We also show that our sensor and activity embedding approach can be easily scaled to other models, including pure temporal models such as LSTM\cite{LSTM} and Transformer\cite{transformer}. We contribute to the research community by publicly releasing our code, dataset, and experiment logs.\footnote{\url{https://github.com/SuminHan/Traffic-UAGCRNTF}}


    
    


    

\section{Related Work}

\subsection{Traffic prediction}

Recent advancements in traffic prediction have prominently relied on the remarkable performance of spatiotemporal neural networks. 
However, the effective utilization of sensor proximity matrices remains an ongoing discussion.
Baseline works which are DCRNN\cite{DCRNN} and STGCN \cite{STGCN} computes sensor proximity matrix from distance with Gaussian filter, and perform graph convolution with the temporal model to facilitate spatiotemporal learning. 
More recently, various researchers have proposed a model architecture that self-trains the sensor similarity during training, mainly based on the attention mechanism. 
ASTGCNN \cite{ASTGCNNTF} uses spatial attention matrix to adjust the weights of the adjacency matrix dynamically and Graph-WaveNet\cite{GraphWavenet} learns the adjacency matrix in an end-to-end manner. While GMAN\cite{GMAN} leverages spatiotemporal embedding based on Node2vec\cite{node2vec} for attention blocks, they still allow a model to train spatial similarity during the spatial attention. STEP\cite{STEP} leverages a Gumbel-Softmax-based trainable adjacency matrix, similar to GTS\cite{GTS}, while leveraging pre-trained k-nearest neighbors (kNN) based graph adjacency for additional graph loss.

\subsection{Urban Human Activity}

Activity has been one of the key topics in travel demand analysis.~\cite{schneider2013unravelling, pinjari2011activity}.
Compared to the traditional trip-based travel demand models (OD-based model), activity-based models are based on the principle that travel demand (trip-chain of ODs) is derived from people's daily activity patterns~\cite{ben1998activity}.
This type of model portrays how people plan and schedule their daily travel~\cite{castiglione2015activity}.
This information lies decision-making process of actual travelers and thus the sequence of urban activity should be found to understand travel behaviors.

While travel forecasting models based on trips have been widely employed, there are many criticisms that their lack of attention to the travel activities which reflect spatial and temporal consequences~\cite{kitamura1988evaluation, spear1996new, mcnally2000four}.
One of the dominant ideas underlying the activity-based approach is that human travels are faced with the inseparability and scarce nature of space and time~\cite{neutens2011prism, axhausen1992activity}.
Activity-based models are becoming popular in transportation as it allows us to infer the consequences of travel behavior that could not have been understood in trip-based models~\cite{joubert2020activity}.

\section{Problem Formulation}

We begin by formally defining the problem of spatiotemporal traffic prediction. We introduce a traffic sensor similarity graph $\mathcal{G}=(V,E,\mathbf{A})$, where $V$ represents the set of sensors, $E$ denotes the set of edges representing sensor similarity, and $\mathbf{A}$ represents the adjacency matrix. Hence, $N=|V|$ signifies the number of traffic sensors in the graph. At each time step $t$, the traffic values of the $N$ sensors are represented by $X_t \in \mathbb{R}^N$. Additionally, we consider the frequency of urban human activity at time step $t$, denoted as $H_t \in \mathbb{R}^{K_H}$, where $K_H$ indicates the number of categories for human activity.

Our problem is to learn a function $f$ that predicts the next $Q$ timesteps of traffic values, given a historical sequence of $P$ timesteps of traffic values and $P+Q$ timesteps of estimated human activity frequencies ${H}_{t-P+1, ..., t+Q}$.

\begin{equation}
[X_{t-P+1}, ..., X_{t}; \mathcal{G}, {H}_{t-P+1, ..., t+Q}] \overset{f}{\rightarrow} X_{t+1}, ..., X_{t+Q}
\end{equation}

\section{Methodology}

In this section, we present the methodology employed to address the traffic prediction problem. Our approach is composed of three key contributions: refined proximity graph construction, spatial sensor embedding, and urban activity embedding. These contributions form the foundation for our model architectures, namely UA-GCRN and UA-GCTransformer, which incorporate the Graph Convolutional Recurrent Network (GCRN) and attention-based Graph Transformer, respectively, as illustrated in Fig.\ref{fig:uagcrn} and\ref{fig:uagctransformer}. The term "UA" denotes the combination of sensor embedding (SE) and activity embedding (AE) added to the input of the encoder and decoder, to distinguish our application.

\subsection{Graph Construction}

\subsubsection{Travel path generation} \label{sec:travel-path-generation}

\begin{figure}[t!]
\includegraphics[width=\columnwidth]{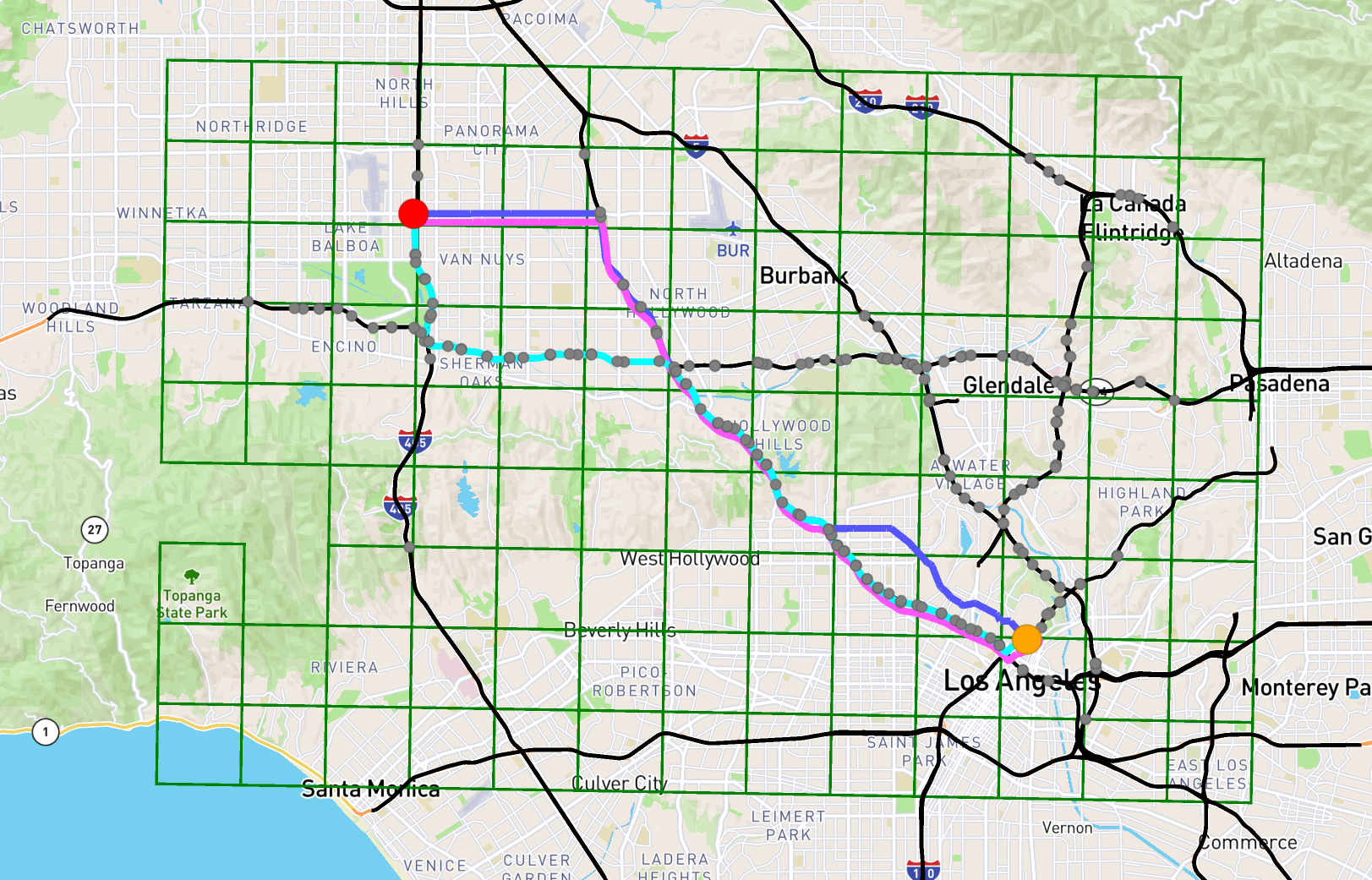}
\vspace{-5mm}
\caption{The A* algorithm is utilized to generate travel paths between the origin (red) and the destination (orange), sampled from a grid pair, with different costs of using the freeway: the ideal shortest path (blue), and paths that make greater use of the freeway (pink, cyan). The gray markers represent the traffic sensors, which do not necessarily appear on the generated travel paths (In METR-LA dataset).}
\label{fig:astar-motorway}
\vspace{-0.5cm}
\end{figure}

We partition the region in which traffic sensors are located into a grid of size $N_H^{(\text{Grid})} \times N_W^{(\text{Grid})}$. To generate plausible paths for each pair of grid regions, we employ the A* algorithm \cite{hart1968formal}, resulting in a set of paths, $\mathcal{M}^{(Gen)}$. The A* algorithm efficiently explores the search space by combining uniform cost search and greedy best-first search. It selects the edge with the minimum cost as it progresses. In the A* algorithm, the cost of the next step movement is determined by the distance of the road, enabling the identification of the shortest path. By applying a coefficient to the cost of freeways during path generation, less than 1, we can obtain multiple paths with varying levels of freeway usage (see Fig.\ref{fig:astar-motorway}).

\subsubsection{Sensor Adjacency Matrix Construction}

Firstly, we define the distance between two sensors, $v_i$ and $v_j$, taking into account the road direction that can be traveled by car. Here, $i, j \in {1, ..., N}$ represent the sensor indices. In our research, traffic sensors are installed on one-way freeways where sensors can be reached in consecutive sequence. As a result, the distance matrix is directed, implying that $\text{dist}(v_i, v_j) \neq \text{dist}(v_j, v_i)$.

To construct the similarity matrix, we apply a Gaussian filter to the distance values. The distance-based proximity matrix between sensors $v_i$ and $v_j$ is computed as $A^{(D)}_{ij} = \exp\left(-\frac{{\text{dist}(v_i, v_j)^2}}{{\sigma^2}}\right) $ if $\text{dist}(v_i, v_j) < \kappa$ else 0.
While previous works \cite{DCRNN,STGCN} leveraged standard deviation
\footnote{The measured standard deviations of distances are 4.97 miles (METR-LA), 3.93 miles (PEMS-BAY), and 6.92 miles (PEMSD7) respectively. However, it is important to note that the specific $\sigma$ value can significantly fluctuate depending on the measured distances between sensors, which can potentially challenge the previous approach.} 
for $\sigma$ and encountered ambiguity in selecting $\kappa$, we consider the specific characteristics of the traffic data. Taking into account the average traffic speed of approximately 60 mph (see Tab.~\ref{tab:datastats}) and an average distance traveled of 5 miles every 5 minutes (1 time-step), we set $\sigma$ to 5 miles. 
Furthermore, we choose $\kappa$ to be 80 miles, representing the maximum distance that can be covered within one hour (12 time-steps), which aligns with the observed maximum speed.

Next, to incorporate generated travel trajectories, we calculate the co-occurrence similarity\cite{cooccur} matrix $A^{(S)}_{ij}$, which measures the likelihood of paths between sensor nodes as follows:

\begin{equation}
A^{(S)}_{ij} = \frac{\text{\# paths $v_i,v_j$ co-appear in $\mathcal{M}^{(Gen)}$}}{{\sqrt{\text{\# paths $v_i$ appears} \times \text{\# paths $v_j$ appears in $\mathcal{M}^{(Gen)}$}}}}
\end{equation}

To obtain the final adjacency matrix to be used for graph convolution, we apply element-wise multiplication of the distance matrix and the co-occurrence matrix as $\mathbf{A} = \mathbf{A}^{(D)} \odot \mathbf{A}^{(S)}$.

\begin{figure}[t!]
  \includegraphics[width=\columnwidth]{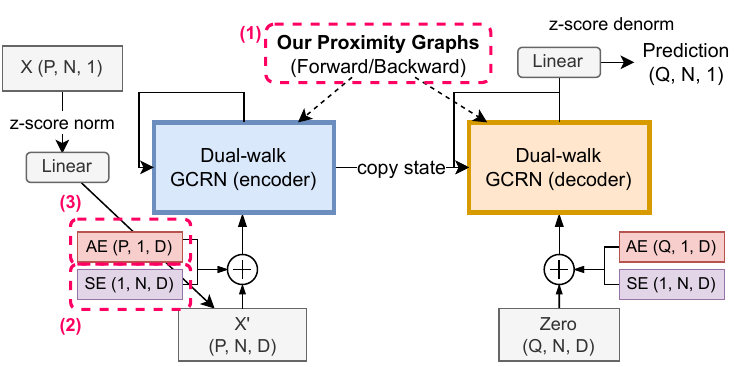}
  \caption{Model Architecture (UA-GCRN)}
  \Description{Simulated}
  \label{fig:uagcrn}
  \vspace{-2mm}
\end{figure}

\begin{figure}[t!]
\vspace{-2mm}
  \includegraphics[width=\columnwidth]{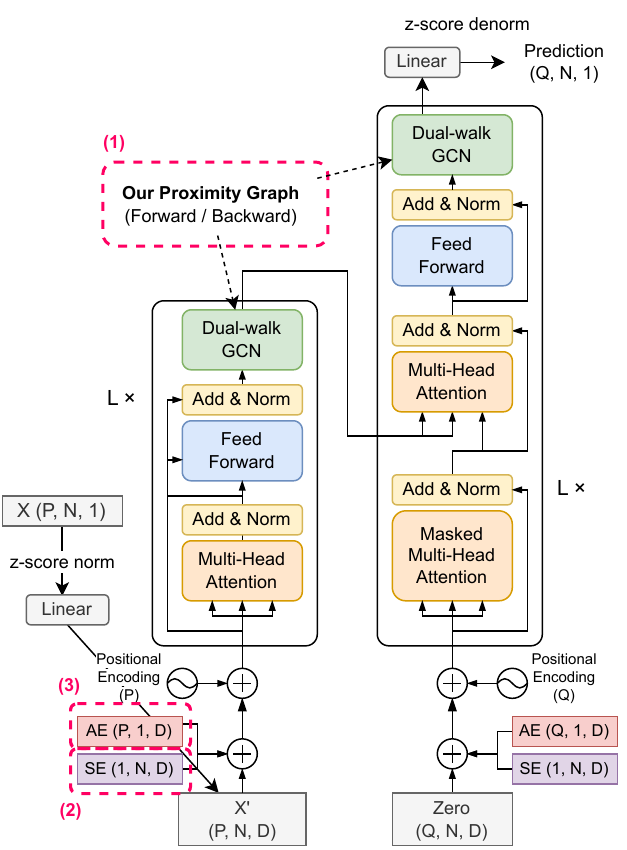}
  \caption{Model Architecture (UA-GCTransformer)}
  \Description{Simulated}
  \label{fig:uagctransformer}
\end{figure}

\begin{figure}[t!]
  \includegraphics[width=\columnwidth]{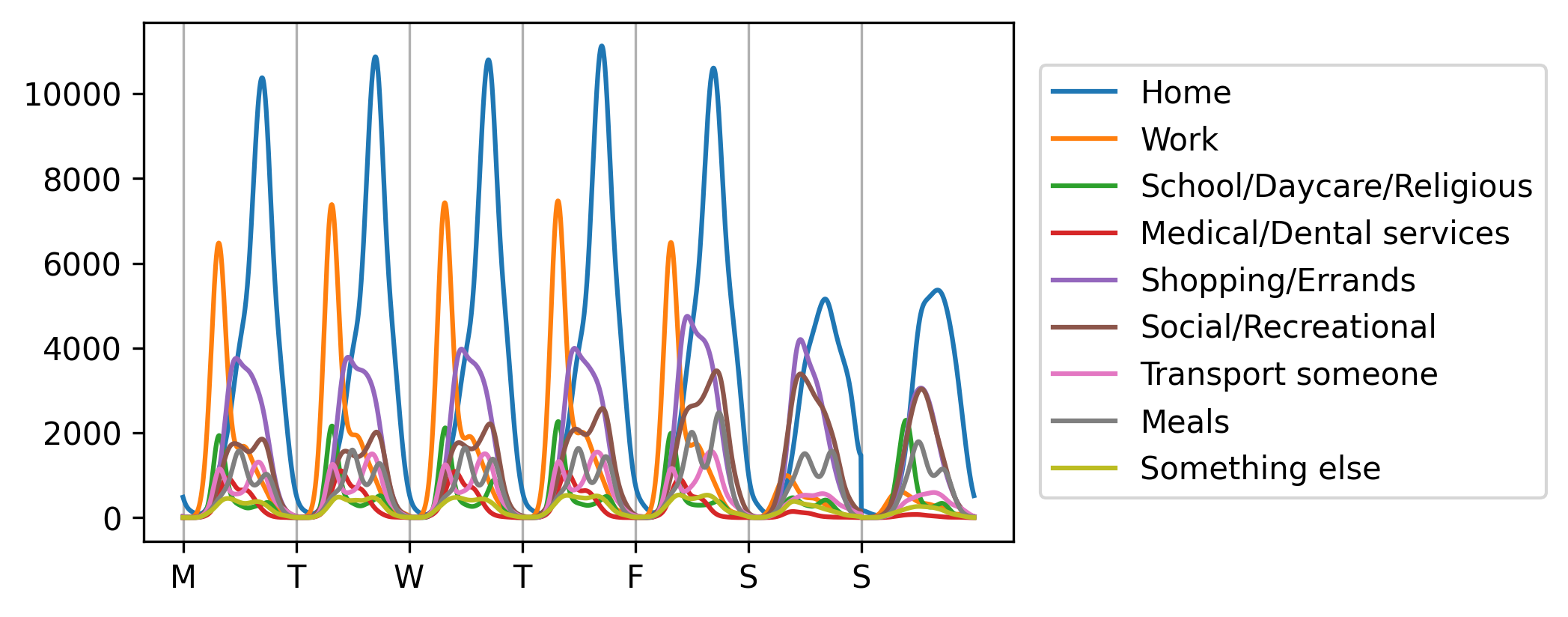}
  \caption{Urban human activity frequencies from the National Household Travel Survey for Activity Embedding.}
  \label{fig:activity-embedding}
  \vspace{-.5cm}
\end{figure}

\subsection{Sensor Embedding}

Each traffic sensor is situated within a unique built environment, resulting in distinct meanings in the actual traffic speed values. However, obtaining reliable sensor metadata to understand these variations is currently limited. To overcome this challenge, we adopt a similar strategy as described in \cite{ASTGCNNTF}, which involves the use of D-dimensional sensor embeddings (SE) generated through one-hot encoding of the $N$ sensors. By incorporating these sensor embeddings, into the input of the encoder and decoder of our models, we can account for the individual characteristics of each sensor.

\subsection{Activity Embedding}

Urban human activity, driven by diverse travel purposes, significantly contributes to traffic congestion \cite{bowman2001activity, bhat2004comprehensive}. To capture the temporal variations in human activity, we construct the activity frequency based on a weekly pattern derived from the National Household Travel Survey \cite{nths-2017}, as depicted in Fig.~\ref{fig:activity-embedding}. This allows us to create a representation $H_{t-P+1, ..., t+Q} \in \mathbb{R}^{K_H}$ that captures the estimated human activities for households at timestamp $t-P+1, ..., t+Q$. Subsequently, we first normalize the activity frequency with standard deviation, and is transformed into a D-dimensional activity embedding using a two-stacked dense layer followed by a normalization layer. To incorporate activity embedding into our models, we including $\text{AE}_{t-P+1, ..., t}$ to the input for the encoder, and  $\text{AE}_{t+1, ..., t+Q}$ to the input for the decoder by addition, along with sensor embeddings. This allows our models to leverage the contextual activity information in both the encoding and decoding stages.

\subsection{Deep Neural Network}

In this section, we present UA-GCRN and UA-GCTransformer as our fundamental models that embody our proposed approach. However, variations of our models, such as UA-LSTM and UA-Transformer, can be implemented without utilizing graph convolution while still considering sensor and activity embedding. To leverage the constructed graph, we introduce dual-walk graph convolution. Additionally, we explore the application of dual-walk graph convolution in two different temporal deep learning methods: recurrent neural network (RNN) and Transformer.

\subsubsection{Dual-walk Graph Convolution}

We utilize a dual-walk graph convolution approach that combines both diffusion and reverse processes. This involves performing a multi-graph convolution using forward walk, backward walk, and the identity matrix.
The motivation behind employing dual-walk convolution is to address traffic congestion that can occur in both directions due to traffic waves \cite{daganzo1994cell}.
The dual-walk graph convolution is equivalent to a single-step dual-walk diffusion convolution, which was initially proposed in \cite{DCRNN}, expressed as follows:

\begin{equation}
g_{\theta \mathcal{G}} Z^{(l+1)} = [\theta_{1} (\mathbf{D}_{out}^{-1}\mathbf{A}) + \theta_{2}(\mathbf{D}_{in}^{-1}\mathbf{A}^T) + \theta_{0} (I)] Z^{(l)}
\end{equation}

\noindent Here, $\theta_0, \theta_1, \theta_2$ represent trainable variables, and $\mathbf{D}_{out}$ and $\mathbf{D}_{in}$ are out-degree and in-degree diagonal matrices of $\mathbf{A}$, respectively. Additionally, we can efficiently perform sparse-matrix computations as our adjacency matrix is sparse.

\subsubsection{Graph Convolutional Recurrent Network}

We apply dual-walk graph convolutional on GCRN \cite{GCRN,DCRNN} as illustrated in Fig.~\ref{fig:uagcrn}. Our UAGCRN module is equivalent to single-step dual-walk diffusion of DCRNN with sensor and activity embedding. 
The reason for employing a single-step instead of a multi-step approach, is due to the incorporation of well-engineered sensor connectivity within our graph structure, rendering the multiple diffusion unnecessary. This is experimentally demonstrated in Sec.~\ref{sec:K-diffusion-bad}. Moreover, a single GCRN module still can accumulate graph convolution of each time-step to enable multi-step prediction.

\subsubsection{Graph Convolutional Transformer}

The Transformer architecture \cite{transformer} has achieved remarkable performance in language modeling tasks, leading to various attempts to adapt its structure for other domains. However, recent studies have relied on learnable positional encoding or learnable graph computation techniques \cite{GMAN, TrafficTransformer, STEP, PDFormer}. Interestingly, no existing model has successfully demonstrated the effectiveness of a basic Transformer on traffic prediction, without graph self-learning or modification in positional encodings. In our UA-GCTransformer model, we adopt the original Transformer architecture and utilize sinusoidal positional encoding to distinguish input and output sequence orders as Fig.~\ref{fig:uagctransformer}. Additionally, we incorporate dual-walk graph convolution in each encoder and decoder layer, similarly in \cite{ASTGCNNTF}. This approach explores the potential of language modeling while leveraging the power of graph convolution.

\begin{table}[t]
\caption{Data statistics (B.C.: Normalized Betweenness Centrality). *PEMSD7 only contains weekdays.}
\vspace{-.2cm}
\begin{tabular}{c|ccc}
\toprule
                   & METR-LA        & PEMS-BAY      & PEMSD7*         \\ \midrule
\# sensors ($N$)         & 207            & 325           & 228            \\
Mean (mph)  & 54 ($\pm$20)  &  62 ($\pm$10)    &  59 ($\pm$13)    \\
Data size & 34,249 & 52,093 & 12,652 \\ 
Start time & Mar/1/2012 & Jan/1/2017 & May/1/2012 \\ 
End time & Jun/30/2012 & May/31/2017 & June/30/2012 \\ \midrule
\# OSM roads       & 75,046        & 36,987        & 122,201               \\
$N_{H}^{(\text{Grid})} \times N_{W}^{(\text{Grid})}$        & 9×13 (2mi.)        & 9×9 (2mi.)       & 8×12 (3mi.)        \\ 
$|\mathcal{M}^{(Gen)}|$ & 105,361         & 46,205      &  66,510              \\
Legacy Adj. NNZ   & 1,722 (4.0\%)  & 2,694 (2.6\%) & 8,100 (15.6\%) \\
Our Adj. NNZ      & 8,575 (20.\%) & 12,628 (12.0\%) & 7,135 (13.7\%) \\
Mean B.C. Ours & $3.04 \times 10^{-3}$ & $2.48 \times 10^{-3}$ &  $3.32 \times 10^{-3}$ 
\\ \bottomrule

\end{tabular}
\label{tab:datastats}
\end{table}

\begin{figure}
  \includegraphics[width=\columnwidth]{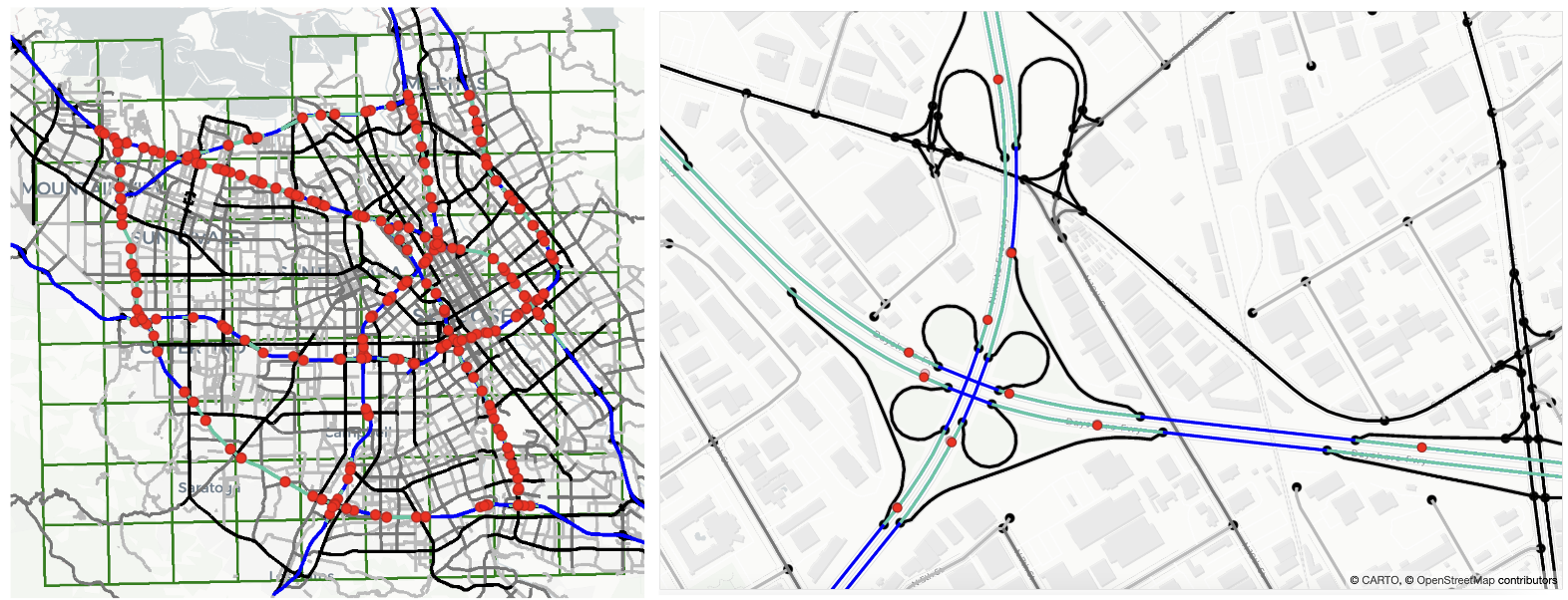}
  \caption{Traffic sensors (red markers) along with OSM freeways (blue paths) and the corresponding freeways where the traffic sensors are located (green paths) in PEMS-BAY. The partitioned grid is also represented with dark green squares.}
  \label{fig:divide-region}
\end{figure}

\section{Experimental Setting}

\subsection{Data Description and Preprocessing}

We provide a description of the datasets used in our study and the corresponding preprocessing steps. Table \ref{tab:datastats} presents the statistics of datasets, including the number of nonzero weights (NNZ) in the adjacency matrix, and the mean betweenness centrality of our adjacency graph.

\subsubsection{Traffic Datasets}
We utilize three well-known traffic datasets: METR-LA \cite{DCRNN}, PEMS-BAY \cite{DCRNN}, and PEMSD7 \cite{STGCN}. These datasets contain information about traffic speeds recorded by sensors, as well as the original sensor adjacency matrix provided by the authors.

\subsubsection{Open Street Map (OSM) Dataset}
To accurately match the sensor locations to the corresponding roads, we leverage Open Street Map\cite{OpenStreetMap} data for the regions covered by the METR-LA, PEMS-BAY, and PEMSD7 datasets. We observed that the locations of some sensors do not align precisely with the OSM roads. In such cases, we updated the latitude, longitude, and freeway details of those sensors using the Caltrans Performance Measurement System (PeMS) \cite{PEMSWebpage}.

\subsubsection{Urban Activity Dataset}
To incorporate urban human activity information, we extracted data from the National Household Travel Survey~\cite{nths-2017}. This dataset contains 828,438 travel surveys that include information about travel start and end times, as well as the mode of transportation (including car usage). We constructed an activity frequency histogram with a 5-minute resolution and smoothed pattern with gaussian filter (sigma=2), and used it as an input for activity embedding in our model. The number of activity categories is $K_H=9$ and described in Fig.~\ref{fig:activity-embedding}.

\subsubsection{Travel Path Generation}
\begin{figure}[t]
  \includegraphics[width=\columnwidth]{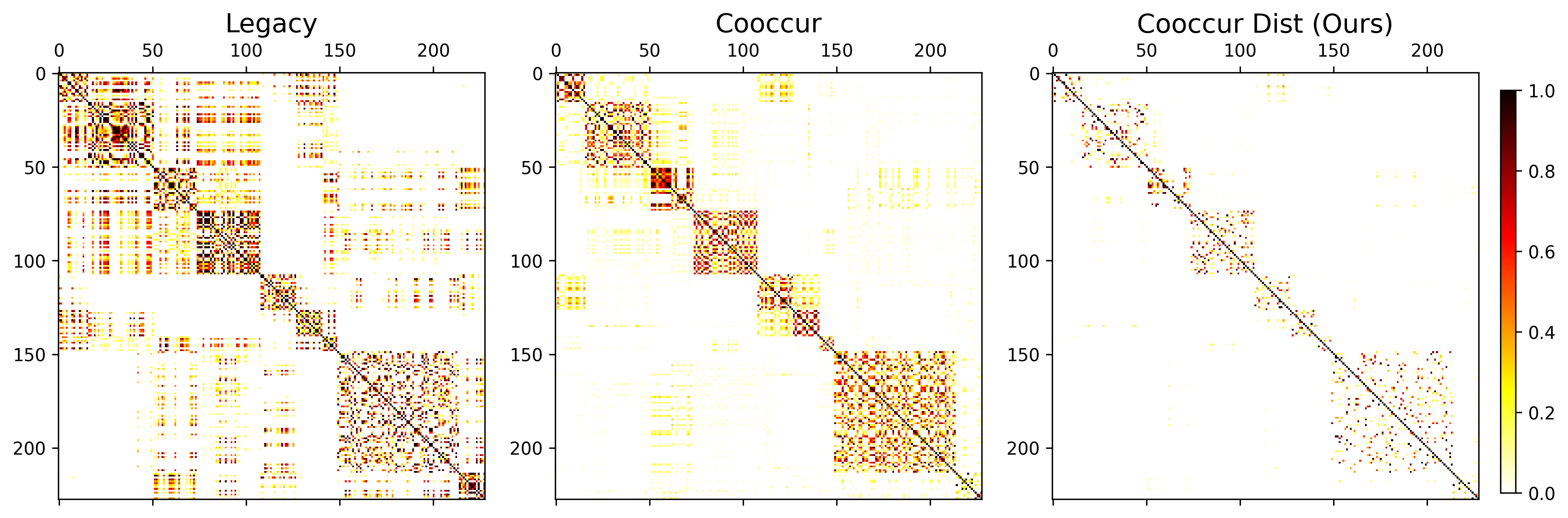}
  \caption{Adjacency Matrix Visualization: Legacy, Co-Occurrence Similarity, and Final Graph (PEMSD7).}
  \label{fig:adjacency-viz}
  \vspace{-.5cm}
  
\end{figure}

We conduct the travel path generation as illustrated in Sec.~\ref{sec:travel-path-generation}. We partition the area around the sensors into square grids of 2-3 miles, including padding, resulting in $N_{H}^{(\text{Grid})} \times N_{W}^{(\text{Grid})}$ grids. For $(N_{H}^{(\text{Grid})} \times N_{W}^{(\text{Grid})}) ^ 2$ pairs of grids, we attempt to generate a travel path using the A* algorithm. We perform this process 5 times with 3 different freeway costs (1.0, 0.9, 0.8 multiplied to freeway road length) for each grid pair as described in Fig.~\ref{fig:astar-motorway}, resulting in a maximum of 15 roads being created\footnote{Note that there can be cases where a path is not established.}. As a result, we can generate $\mathcal{M}^{(Gen)}$ travel paths to construct our co-occurrence and distance-based adjacency matrix. As a example, adjacency matrix of PEMSD7 is visualized in Fig.~\ref{fig:adjacency-viz}.

\subsection{Evaluation Setup}
In our experiments, we evaluate MAE (Mean Absolute Error), RMSE (Root Mean Square Error), and MAPE (Mean Absolute Percentage Error) at 3, 6, and last (12 in METR-LA, PEMS-BAY, 9 in PEMSD7) step of prediction.

We utilized the following parameter settings: a batch size of 32, a hidden embedding dimension ($D$) of 64\footnote{Dimension of 64 is a frequently employed hyperparameter choice in both DCRNN and GMAN architectures, for fair comparison.}, and the Adam optimizer with an initial learning rate of 0.01. We employed a patience of 5 for early stopping and reduced the learning rate to 1/10 after 2 trials. For the Transformer models, we employed 8 attention heads, a key dimension of 8, a total dimension of 64, and stacked 3 layers.

\section{Results}

\begin{table*}[t]
\centering
\caption{Forecasting error in METR-LA, PEMS-BAY, PEMSD7 datasets. $\dagger$ represents the model leveraging our co-occurence and distance based adjacency matrix. $^\ast$ represents the model self trains the sensor similarity. Best and second best results are represented as BOLD and \underline{underline}.}
\label{tab:result-overall}
\begin{tabular}{c|c|c|ccc|cc|ccccccc|cc}
\toprule
 &  &  & \multicolumn{5}{c|}{Temporal Only} & \multicolumn{9}{c}{Spatiotemporal} \\ 
 &  & Metric & \rotatebox[origin=c]{90}{LastRepeat} & \rotatebox[origin=c]{90}{LSTM} & \rotatebox[origin=c]{90}{TF} & \rotatebox[origin=c]{90}{UA-LSTM} & \rotatebox[origin=c]{90}{UA-TF} & \rotatebox[origin=c]{90}{DCRNN} & \rotatebox[origin=c]{90}{DCRNN$\dagger$} & \rotatebox[origin=c]{90}{GTS$^\ast$\cite{STEP}} & \rotatebox[origin=c]{90}{STGCN} & \rotatebox[origin=c]{90}{GWNet$^\ast$} & \rotatebox[origin=c]{90}{GMAN$^\ast$} & \rotatebox[origin=c]{90}{STEP$^\ast$}  & \rotatebox[origin=c]{90}{\textbf{UAGCRN$\dagger$}} & \rotatebox[origin=c]{90}{UAGCTF$\dagger$} \\ \midrule

\multirow{9}{*}{\rotatebox[origin=c]{90}{\textit{METR-LA}}}  & \multirow{3}{*}{\rotatebox[origin=c]{90}{15 min}} & MAE & 4.02 & 3.09 & 3.07 & 2.82 & 2.81 & 2.77 & 2.67 & 2.67 & 2.88 & 2.69 & 2.77 & \textbf{2.61} & 2.64 & \underline{2.63}\\
 &  & RMSE & 8.69 & 6.10 & 6.09 & 5.56 & 5.58 & 5.38 & 5.21 & 5.27 & 5.74 & 5.15 & 5.48 & \textbf{4.98 }& 5.09 & \underline{5.07}\\
 &  & MAPE & 9.4\% & 8.2\% & 8.1\% & 7.5\% & 7.6\% & 7.30\% & 6.89\% & 7.21\% & 7.62\% & 6.90\% & 7.25\% & \textbf{6.60\%} & 6.77\% & \underline{6.71\%}\\
\cline{2-17}
 & \multirow{3}{*}{\rotatebox[origin=c]{90}{30 min}} & MAE & 5.09 & 3.79 & 3.76 & 3.19 & 3.18 & 3.15 & 3.06 & 3.04 & 3.47 & 3.07 & 3.07 & \textbf{2.96} & 2.97 & \underline{2.96}\\
 &  & RMSE & 11.13 & 7.66 & 7.65 & 6.59 & 6.61 & 6.45 & 6.30 & 6.25 & 7.24 & 6.22 & 6.34 & \textbf{5.97} & 6.08 & \underline{6.04}\\
 &  & MAPE & 12.2\% & 10.7\% & 10.6\% & 9.1\% & 9.1\% & 8.80\% & 8.38\% & 8.41\% & 9.57\% & 8.37\% & 8.35\% & \textbf{7.96\%} & 8.10\% & \underline{8.08\%}\\
\cline{2-17}
 & \multirow{3}{*}{\rotatebox[origin=c]{90}{60 min}} & MAE & 6.80 & 4.90 & 4.88 & 3.56 & 3.54 & 3.60 & 3.56 & 3.46 & 4.59 & 3.53 & 3.40 & 3.37 & \underline{3.35} & \textbf{3.34}\\
 &  & RMSE & 14.21 & 9.68 & 9.67 & 7.55 & 7.52 & 7.59 & 7.52 & 7.31 & 9.40 & 7.37 & 7.21 & \textbf{6.99} & 7.12 & \underline{7.02}\\
 &  & MAPE & 16.7\% & 14.9\% & 14.8\% & 10.6\% & 10.7\% & 10.50\% & 10.15\% & 9.98\% & 12.70\% & 10.01\% & 9.72\% & \textbf{9.61\%} & 9.68\% & \underline{9.65\%}\\
\midrule
\multirow{9}{*}{\rotatebox[origin=c]{90}{\textit{PEMS-BAY}}}  & \multirow{3}{*}{\rotatebox[origin=c]{90}{15 min}} & MAE & 1.60 & 1.45 & 1.45 & 1.32 & 1.33 & 1.38 & \underline{1.29} & 1.34 & 1.36 & 1.30 & 1.34 & \textbf{1.26} & 1.30 & 1.30\\
 &  & RMSE & 3.43 & 3.16 & 3.16 & 2.82 & 2.85 & 2.95 & \textbf{2.72} & 2.83 & 2.96 & 2.74 & 2.82 & \underline{2.73} & 2.73 & 2.76\\
 &  & MAPE & 3.2\% & 3.0\% & 3.0\% & 2.8\% & 2.8\% & 2.90\% & \underline{2.69\%} & 2.82\% & 2.90\% & 2.73\% & 2.81\% & \textbf{2.59\%} & 2.71\% & 2.75\%\\
\cline{2-17}
 & \multirow{3}{*}{\rotatebox[origin=c]{90}{30 min}} & MAE & 2.18 & 1.98 & 1.98 & 1.63 & 1.63 & 1.74 & 1.62 & 1.66 & 1.81 & 1.63 & 1.62 & \textbf{1.55} & \underline{1.61} & 1.61\\
 &  & RMSE & 4.99 & 4.61 & 4.61 & 3.77 & 3.78 & 3.97 & 3.68 & 3.78 & 4.27 & 3.70 & 3.72 & \textbf{3.58} & \underline{3.68} & 3.70\\
 &  & MAPE & 4.7\% & 4.5\% & 4.5\% & 3.7\% & 3.7\% & 3.90\% & 3.62\% & 3.77\% & 4.17\% & 3.67\% & 3.63\% & \textbf{3.43\%} & \underline{3.62\%} & 3.64\%\\
\cline{2-17}
 & \multirow{3}{*}{\rotatebox[origin=c]{90}{60 min}} & MAE & 3.05 & 2.72 & 2.71 & 1.89 & 1.88 & 2.07 & 1.92 & 1.95 & 2.49 & 1.95 & \underline{1.86} & \textbf{1.79} & 1.87 & 1.86\\
 &  & RMSE & 7.01 & 6.28 & 6.27 & 4.41 & 4.40 & 4.74 & 4.45 & 4.43 & 5.69 & 4.52 & \underline{4.32} & \textbf{4.20} & 4.37 & 4.33\\
 &  & MAPE & 6.8\% & 6.8\% & 6.7\% & 4.5\% & 4.4\% & 4.90\% & 4.52\% & 4.58\% & 5.79\% & 4.63\% & \underline{4.31\%} & \textbf{4.18\%} & 4.39\% & 4.36\%\\
\midrule
\multirow{9}{*}{\rotatebox[origin=c]{90}{\textit{PEMSD7}}}  & \multirow{3}{*}{\rotatebox[origin=c]{90}{15 min}} & MAE & 2.49 & 2.35 & 2.37 & 2.13 & 2.13 & 2.21 & 2.10 & 2.21 & 2.25 & 2.31 & 2.30 & 2.09 & \textbf{2.05} & \underline{2.06}\\
 &  & RMSE & 4.65 & 4.48 & 4.51 & 4.03 & 4.12 & 4.21 & 3.98 & 4.16 & 4.04 & 4.44 & 4.39 & 3.99 & \textbf{3.87} & \underline{3.93}\\
 &  & MAPE & 5.7\% & 5.5\% & 5.5\% & 5.1\% & 5.1\% & 5.14\% & 4.91\% & 5.15\% & 5.26\% & 5.41\% & 5.66\% & 5.00\% & \textbf{4.85\%} & \underline{4.86\%}\\
\cline{2-17}
 & \multirow{3}{*}{\rotatebox[origin=c]{90}{30 min}} & MAE & 3.51 & 3.31 & 3.33 & 2.71 & 2.69 & 3.01 & 2.75 & 2.95 & 3.03 & 3.26 & 2.71 & 2.66 & \underline{2.61} & \textbf{2.59}\\
 &  & RMSE & 6.77 & 6.49 & 6.53 & 5.37 & 5.49 & 5.96 & 5.45 & 5.74 & 5.70 & 6.41 & 5.35 & 5.37 & \textbf{5.20} & \underline{5.22}\\
 &  & MAPE & 8.3\% & 8.1\% & 8.1\% & 6.9\% & 6.9\% & 7.43\% & 6.85\% & 7.43\% & 7.33\% & 8.11\% & 6.87\% & 6.80\% & \underline{6.56\%} & \textbf{6.50\%}\\
\cline{2-17}
 & \multirow{3}{*}{\rotatebox[origin=c]{90}{45 min}} & MAE & 4.31 & 4.05 & 4.10 & 3.01 & 2.98 & 3.59 & 3.19 & 3.47 & 3.57 & 4.63 & 2.99 & 2.95 & \underline{2.92} & \textbf{2.90}\\
 &  & RMSE & 8.32 & 7.89 & 7.99 & 6.10 & 6.13 & 7.14 & 6.39 & 6.78 & 6.77 & 8.81 & 5.94 & 6.03 & \textbf{5.90} & \underline{5.91}\\
 &  & MAPE & 10.4\% & 10.3\% & 10.3\% & 7.9\% & 7.8\% & 9.18\% & 8.24\% & 9.06\% & 8.69\% & 12.40\% & 7.70\% & 7.74\% & \underline{7.58\%} & \textbf{7.48\%}\\
 \bottomrule
\end{tabular}
\end{table*}


\begin{figure*}
    \centering
    
    \begin{subfigure}{0.49\textwidth}
        \includegraphics[width=\linewidth]{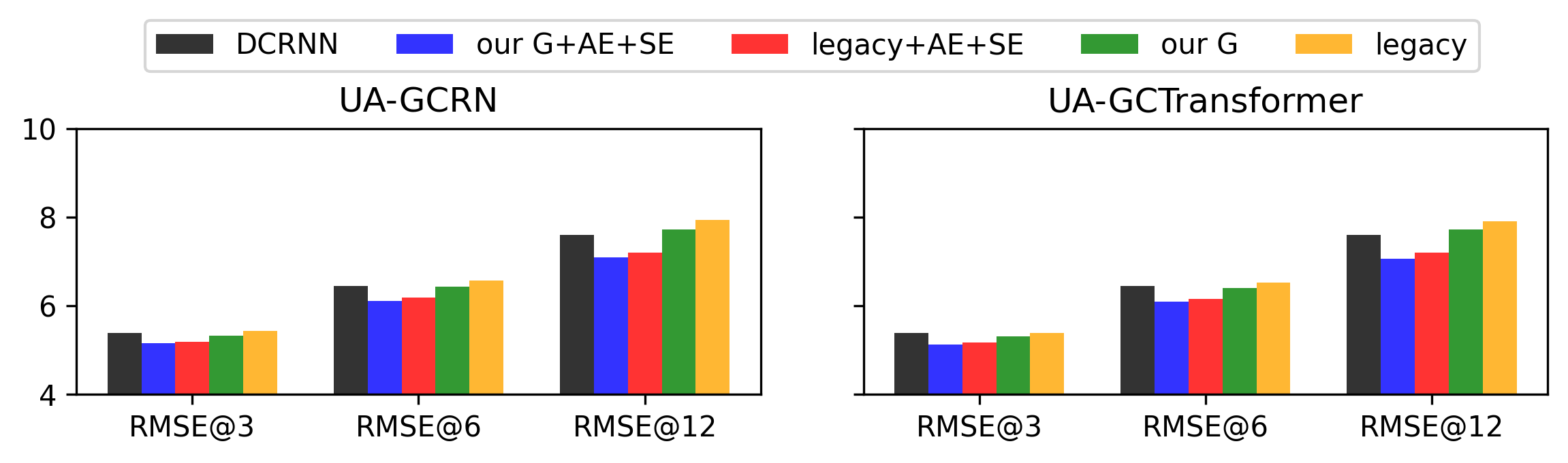}
        \caption{METR-LA (UA-GCRN, UA-GCTransformer)}
        \label{fig:subplot1}
    \end{subfigure}
    \hfill
    \begin{subfigure}{0.49\textwidth}
        \includegraphics[width=\linewidth]{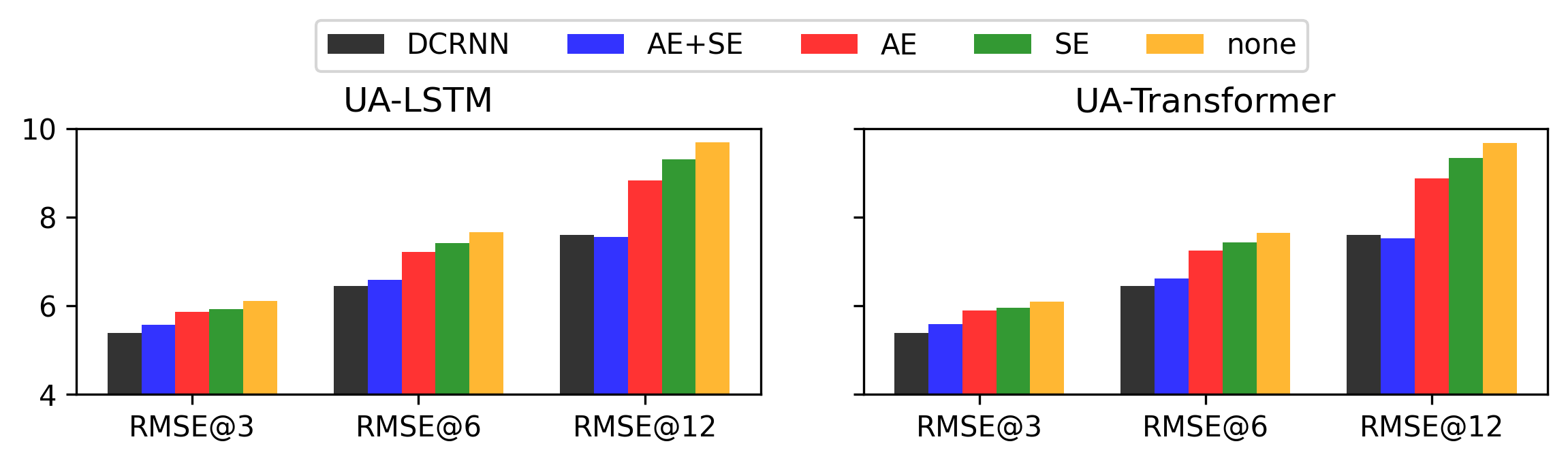}
        \caption{METR-LA (UA-LSTM, UA-Transformer)}
        \label{fig:subplot2}
    \end{subfigure}
    
    
    \begin{subfigure}{0.49\textwidth}
        \includegraphics[width=\linewidth]{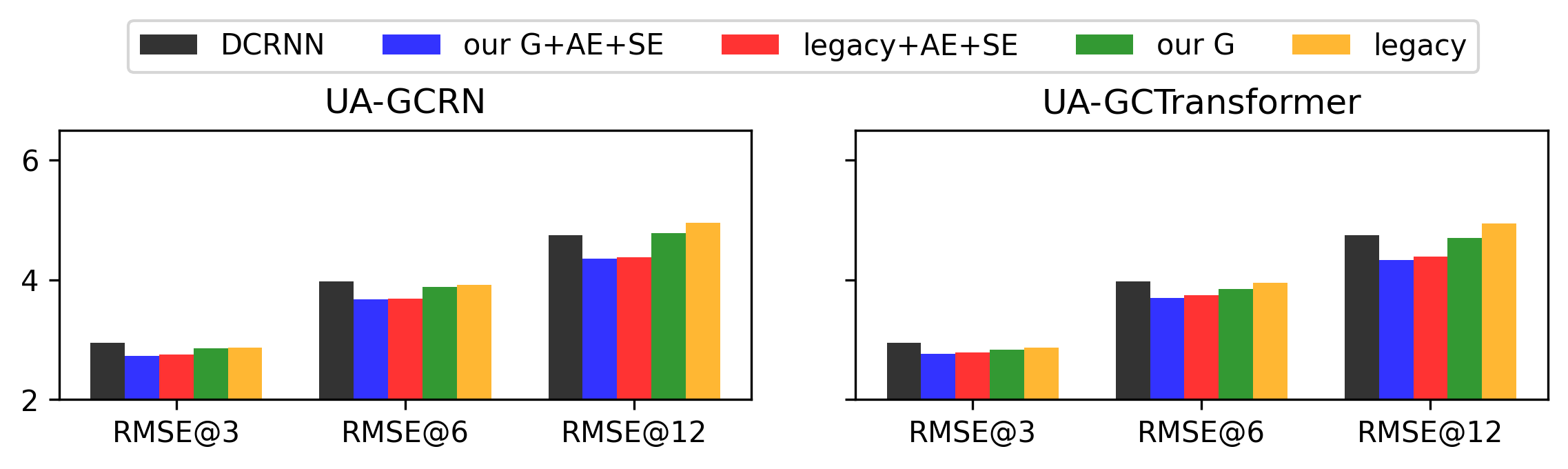}
        \caption{PEMS-BAY (UA-GCRN, UA-GCTransformer)}
        \label{fig:subplot3}
    \end{subfigure}
    \hfill
    \begin{subfigure}{0.49\textwidth}
        \includegraphics[width=\linewidth]{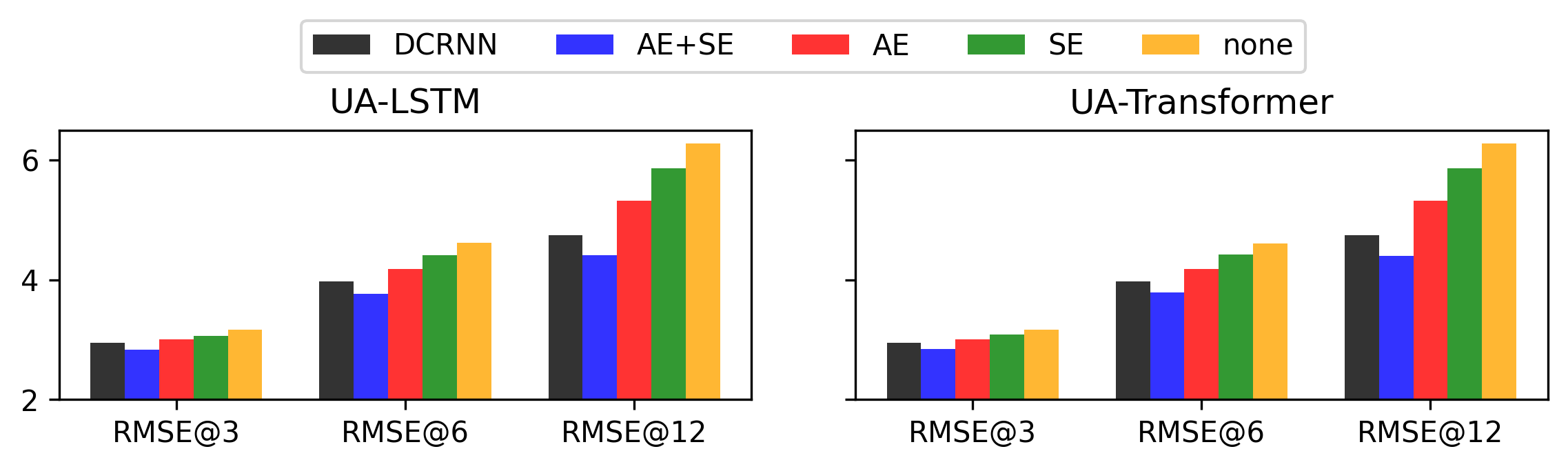}
        \caption{PEMS-BAY (UA-LSTM, UA-Transformer)}
        \label{fig:subplot4}
    \end{subfigure}
    
    
    \begin{subfigure}{0.49\textwidth}
        \includegraphics[width=\linewidth]{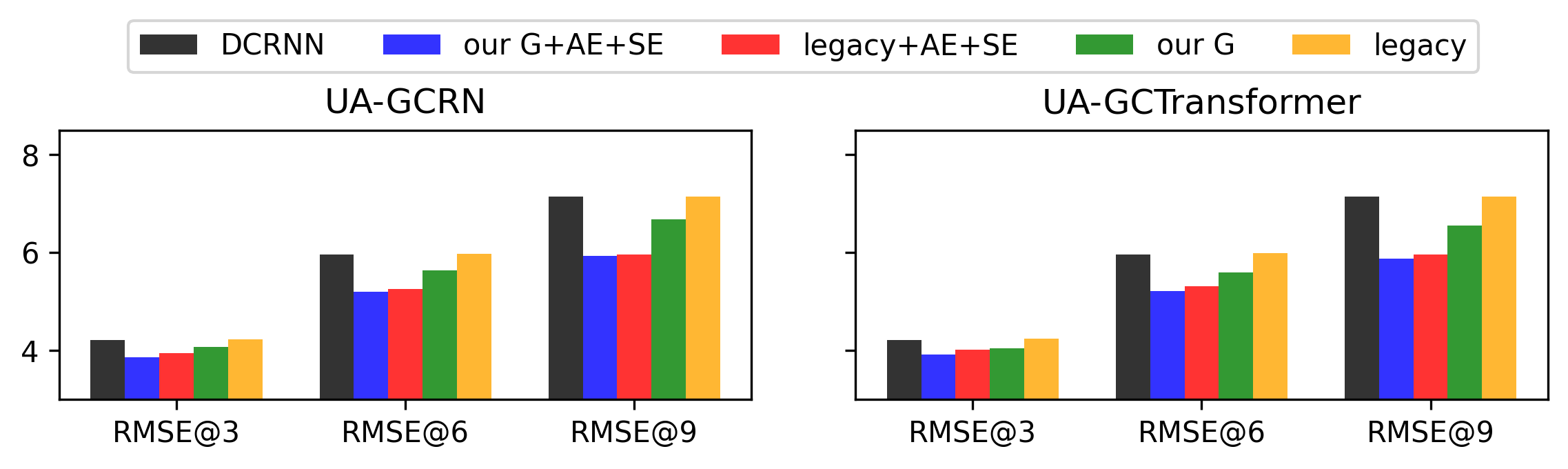}
        \caption{PEMSD7 (UA-GCRN, UA-GCTransformer)}
        \label{fig:subplot5}
    \end{subfigure}
    \hfill
    \begin{subfigure}{0.49\textwidth}
        \includegraphics[width=\linewidth]{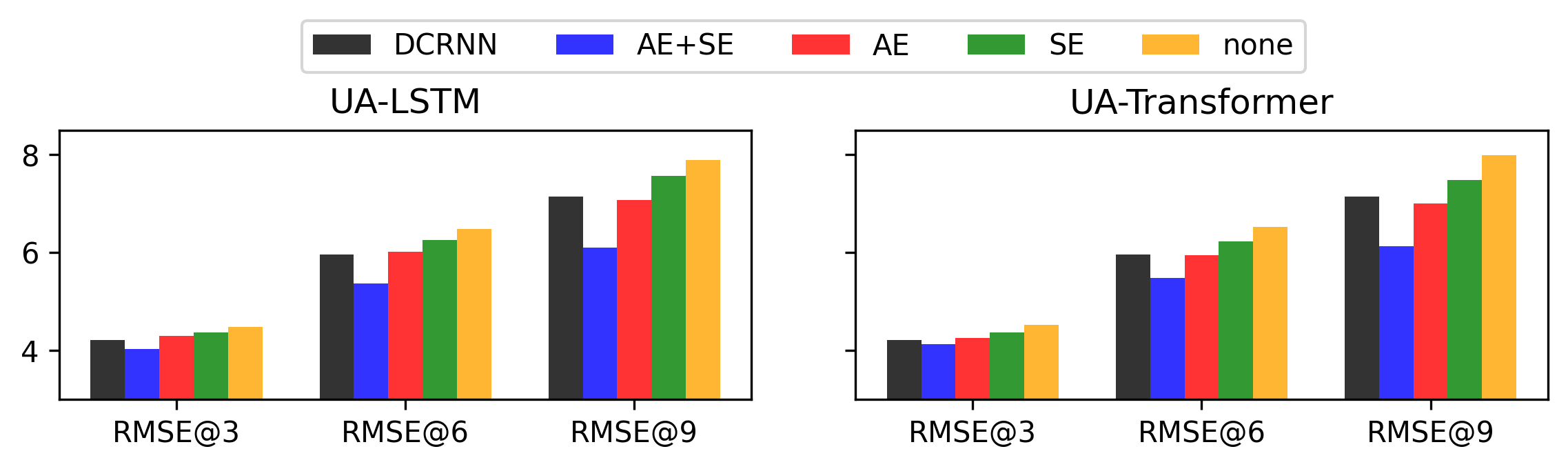}
        \caption{PEMSD7 (UA-LSTM, UA-Transformer)}
        \label{fig:subplot6}
    \end{subfigure}
    
    \caption{Ablation Test (RMSE) of our modules  -- Our Graph(G), SE, AE.}
    \label{fig:ablation-test}
\end{figure*}

\subsection{Performance Comparison}

We have selected several baselines for comparison with our proposed UAGCRN, UAGCTransformer. The baselines include\footnote{Although we considered Traffic Transfomer\cite{TrafficTransformer} for its state-of-the-art performance, we encountered difficulties in finding a reliable dataset or test logs.  We assume there might be confusion in metrics such as averaging over multiple time steps.}:
Last Repeat, LSTM, Transformer\cite{transformer}, DCRNN\cite{DCRNN}, GTS\cite{GTS}, STGCN\cite{STGCN}, Graph Wavenet (GWNet) \cite{GraphWavenet}, GMAN\cite{GMAN}, STEP\cite{STEP}.

  
  
  
  

  
  
  

Additionally, we also compare our proposed models with two variants: UA-LSTM and UA-Transformer. These variants leverage activity and sensor embeddings but do not incorporate graph convolutions, enabling us to evaluate the impact of graph utilization.

\subsubsection{Forecasting error}

Tab.~\ref{tab:result-overall} presents the results of comparing various baseline models with our proposed models (UAGCRN and UAGCTransformer). On the header of the table, $\dagger$ represents the model leveraging our co-occurrence and distance-based adjacency matrix, and $^\ast$ represents the model self-learns the sensor similarity, which are GTS, GWNet, GMAN, and STEP.

Overall, the proposed models, UAGCRN and UAGCTransformer, consistently outperform the major spatiotemporal baselines across all three datasets and various time intervals. Moreover, these models outperform UALSTM and UATransformer, indicating that incorporating graph convolutions significantly improves the accuracy of traffic forecasting models.

We further evaluate the effectiveness of our graph construction approach by comparing DCRNN, DCRNN$\dagger$, and GTS\footnote{Results from \cite{STEP} due to issues with GTS: \url{https://github.com/chaoshangcs/GTS/issues}}, which share the same architecture. Notably, we observed that DCRNN$\dagger$ outperforms GTS, particularly on the PEMS-BAY and PEMSD7 datasets. The superior performance of DCRNN$\dagger$ can be attributed to the fact that GTS considers all potential sensor connections, often resulting in biased predictions. We also noticed this issue of trainable graph similarity in GWNet, as it exhibits significant errors on the PEMSD7 dataset. The sensor networks in the PEMSD7 dataset have higher betweenness centrality (Tab.~\ref{tab:datastats}), indicating that the graph structure provides more valuable information compared to other datasets. These complexities in sensor similarity may also contribute to the lower performance of the STEP model compared to our approach, as STEP relies on a data-driven approach that may not accurately capture the intricate sensor relationships.

Moreover, our proposed UA approach, which includes SE and AE components, significantly improves the performance of purely temporal models (LSTM and TF). UA-LSTM and UA-Transformer surpass other spatiotemporal baselines such as DCRNN, GTS, STGCN, GWNet, and GMAN on the PEMS-BAY and PEMSD7 datasets. 
This observation suggests that by incorporating $(P+Q) \times N$ types of inputs, which include both sensor index and urban activity context, our models are able to distinguish between different sensor inputs and capture distinct activity contexts. This comprehensive input representation empowers our models to generate accurate predictions for multi-step traffic forecasting tasks.

However, we observe limited improvement of Transformer over LSTM and UAGCTransformer over UAGCRN. This can be attributed to the fact that traffic prediction involves relatively short time-series steps, unlike in the case of large language models, where Transformers excel. Consequently, RNN models continue to perform well in this domain.

Although our proposed models, UAGCRN$\dagger$ and UAGCTF$\dagger$, are outperformed by STEP on the METR-LA and PEMS-BAY datasets, STEP utilizes very long patches of input (e.g. $P=228\times7$) and complex transformer architecture which allows it to capture more complex and intricate patterns. This approach results in a heavier model that can handle larger contextual information, as it shows better performance in longer timesteps. Despite the performance difference, our models still demonstrate potential for improvement, which will be explained in Sec.~\ref{sec:TE-better-AE}. On the other hand, we believe that studying STEP's architecture and mechanisms can provide valuable insights to advance the state-of-the-art in related models.


\subsubsection{Computational Cost}

We conducted a comparison of the computational cost for each model in their default settings in Tab.~\ref{tab:computation-cost}\footnote{Tab.~\ref{tab:result-overall} is based solely on the original author's implementation, while Tab.~\ref{tab:computation-cost} is intended for evaluating computational time under same learning framework (TensorFlow2) and GPU (RTX3090), batch size, and early stopping condition.}. In order to ensure a fair comparison we leverage the default settings of each model such as DCRNN with 3 diffusion steps, GMAN with $L=5$. We were unable to precisely measure the computational cost of STEP\cite{STEP} under the same environment. However, during our experiments of STEP with the PEMSD7 dataset (2.7 times smaller than METR-LA), each epoch took approximately 3.5 minutes to train using 3 TITAN RTX GPUs. The total training time was approximately 5 hours.
The result shows that UAGCRN outperforms other models in terms of computational cost and training time.

\begin{table}[h]
\caption{Computational cost of METR-LA under same environment. The number of stacks are $L=5$ in GMAN and $L=3$ in UAGCTF$\dagger$, while DCRNN, UAGCRN$\dagger$ do not have stacked architecture ($L=1$).}
\vspace{-2mm}
\label{tab:computation-cost}
\begin{tabular}{c|cccc}
\toprule
 & DCRNN & GMAN & UAGCRN$\dagger$ & UAGCTF$\dagger$ \\ \midrule
\# Params & 353,025 & 714,049 & \textbf{174,401} & 842,177 \\
Train (m:s/ep.) & 2:35 & 4:39 & \textbf{42s} &  4:26  \\ 
Total Epochs & 26 & 19 & 24 & 17 \\
Total train time & 1:12:03 & 1:34:41 & \textbf{0:18:36} & 1:21:04 \\
\bottomrule
\end{tabular} 
\end{table}

\subsection{Ablation Study}

\subsubsection{Effectness of the individual module (Graph, AE, SE)}

The results of the ablation test for each module are presented in Fig.~\ref{fig:ablation-test}. Specifically, Fig.~\ref{fig:subplot1},\ref{fig:subplot3},\ref{fig:subplot5} demonstrate the performance improvement of UA-GCRN and UA-GCTransformer, when using our graph compared to the legacy graph. These results indicate that our graph contains more traffic-related knowledge regarding sensor correlation. Although the enhancement looks marginal when both the SE and AE modules are given under the same conditions, it still highlights the potential for performance improvement in less common situations.

Furthermore, Fig.~\ref{fig:subplot2},\ref{fig:subplot3},\ref{fig:subplot6} showcase the effectiveness of each SE and AE module on temporal-only models, specifically UA-LSTM and UA-Transformer. The performance improves as these modules are integrated. Notably, the impact of the AE module is more significant than the SE module, likely due to traffic patterns exhibiting a stronger correlation with human activity. Additionally, incorporating the SE module to account for sensor spatial heterogeneity further enhances performance. When both SE and AE modules are integrated, the resulting performance surpasses that of DCRNN in the PEMS-BAY and PEMSD7 datasets.

\subsubsection{The number of diffusion steps of DCRNN with our graph} \label{sec:K-diffusion-bad}

\begin{figure}[h]
  \includegraphics[width=\columnwidth]{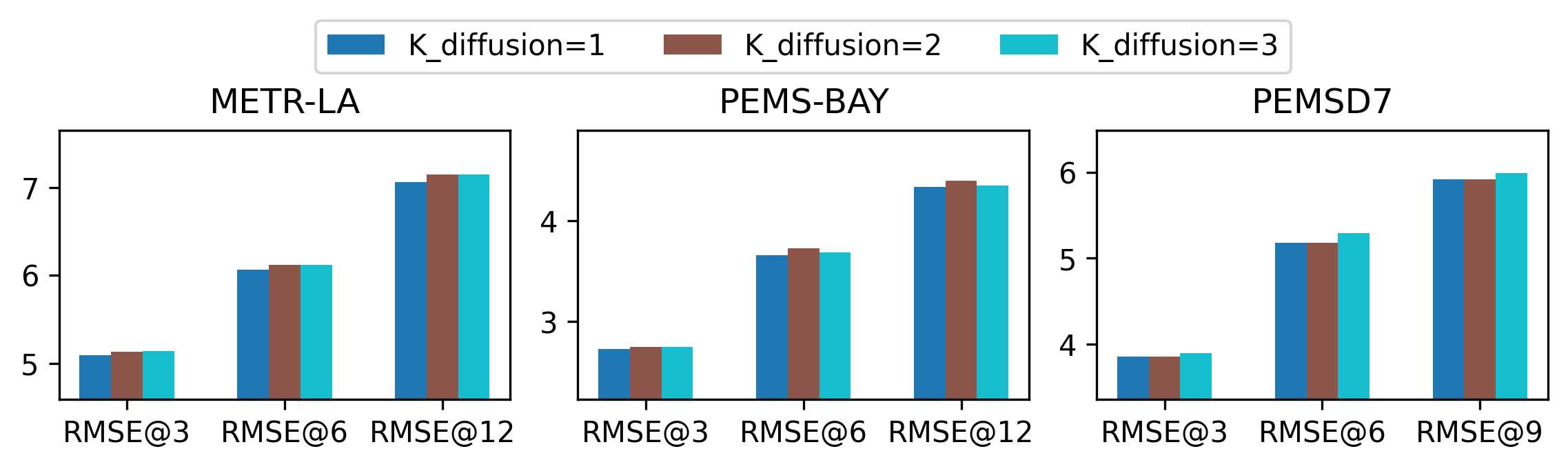}
  \caption{Performance degradation in UADCGRU$\dagger$ as the number of diffusion steps (K) increases.}
  \label{fig:K-diffusion-bad}
  \vspace{-2mm}
\end{figure}

Figure~\ref{fig:K-diffusion-bad} depicts the results of UADCGRU$\dagger$ obtained by applying the SE and AE to the DCGRU model using our graph while modifying the diffusion steps. In contrast to the original findings discussed in \cite{DCRNN} which suggested a demand for approximately 3 diffusion steps, our results show that increasing the graph connectivity information, along with activity and sensor data, can lead to worse performance. We analyze that vehicles do not follow a random-walk pattern, and the vehicle travel pattern is already adequately captured in our constructed graph. 

\subsubsection{Comparison of Timestamp Embedding and Activity Embedding} \label{sec:TE-better-AE}

Various models have employed different approaches to incorporate timestamp information. For example, in DCRNN, the time of day is included as an additional input channel\footnote{Not mentioned in the paper, but in the code: \url{https://github.com/liyaguang/DCRNN}}. In this ablation study, we compared timestamp embedding (TE), which are generated from a vector space $\{0,1\}^{7+12\times24}$ (one-hot 
 concatenation of weekday and time-of-day) and ingested in 2-stacked dense layer with normalization layer to capture weekly and daily periodicity, similar to \cite{GMAN,PDFormer}, with AE.

Tab.~\ref{tab:TE-better-than-AE} shows the comparison results of UAGCRN$\dagger$ with TE or AE, indicating that TE exhibited a slight improvement over AE, performing almost as well as STEP in the METR-LA and PEMS-BAY datasets, and outperforming AE in the PEMSD7 dataset. The performance improvement of the TE over the AE can be attributed to the lack of analysis in localized activity patterns of each city when we estimate human activity frequency which is derived from national surveys.
This aspect suggests that future studies should consider accurately inputting AE, such as localized activity estimation considering demographics and urban function.

On the other hand, relying on one-hot timestamp information results in less explainability due to its discrete nature, unlike the continuous activity information. Additionally, it may pose limitations in scalability when accounting for seasonal effects in long-term datasets, while our datasets are deal with only a few months (Tab.~\ref{tab:datastats}). Nevertheless, we can still take advantage of the AE-based UAGCRN model for its superior explainability.

\begin{table}[t]
\caption{Ablation study of UAGCRN$\dagger$ and UAGCTF$\dagger$ by replacing AE with timestamp embedding (TE). Best and second best results are represented as BOLD and \underline{underline}.}
\vspace{-3mm}
\label{tab:TE-better-than-AE}
\begin{tabular}{c|c|c|cc|cc}
\toprule
 &  & \multirow{2}{*}{STEP} & \multicolumn{2}{c|}{UAGCRN$\dagger$} & \multicolumn{2}{c}{UAGCTF$\dagger$} \\
 &  &  & TE+SE & AE+SE & TE+SE & AE+SE \\ \midrule
\multirow{6}{*}{\rotatebox[origin=c]{90}{METR-LA}} & MAE3 & \textbf{2.61} & \underline{2.62} & 2.64 & 2.63 & 2.63 \\
 & RMSE3 & \textbf{4.98} & \underline{5.00} & 5.09 & 5.09 & 5.07 \\ \cline{2-7}
 & MAE6 & \underline{2.96} & \textbf{2.94} & 2.97 & 2.95 & 2.96 \\
 & RMSE6 & \underline{5.97} & \textbf{5.97} & 6.08 & 6.05 & 6.04 \\ \cline{2-7}
 & MAE12 & 3.37 & \textbf{3.31} & 3.35 & 3.35 & \underline{3.34} \\
 & RMSE12 & \textbf{6.99} & \underline{7.02} & 7.12 & 7.10 & 7.02 \\ \midrule
\multirow{6}{*}{\rotatebox[origin=c]{90}{PEMS-BAY}} & MAE3 & \textbf{1.26} & \underline{1.28} & 1.30 & 1.28 & 1.30 \\
 & RMSE3 & \underline{2.73} & \textbf{2.69} & 2.73 & 2.72 & 2.76 \\ \cline{2-7}
 & MAE6 & \textbf{1.55} & 1.60 & 1.61 & \underline{1.59} & 1.61 \\
 & RMSE6 & \textbf{3.58} & \underline{3.63} & 3.68 & 3.66 & 3.70 \\ \cline{2-7}
 & MAE12 & \textbf{1.79} & 1.88 & 1.87 & 1.86 & \underline{1.86} \\
 & RMSE12 & \textbf{4.20} & 4.38 & 4.37 & 4.37 & \underline{4.33} \\ \midrule
\multirow{6}{*}{\rotatebox[origin=c]{90}{PEMSD7}} & MAE3 & 2.09 & \textbf{2.02} & 2.05 & \underline{2.04} & 2.06 \\
 & RMSE3 & 3.99 & \textbf{3.81} & \underline{3.87} & 3.88 & 3.93 \\ \cline{2-7}
 & MAE6 & 2.66 & \textbf{2.56} & 2.61 & \underline{2.57} & 2.59 \\
 & RMSE6 & 5.37 & \textbf{5.14} & 5.20 & \underline{5.16} & 5.22 \\ \cline{2-7}
 & MAE9 & 2.95 & \textbf{2.88} & 2.92 & \underline{2.89} & 2.90 \\
 & RMSE9 & 6.03 & \textbf{5.88} & 5.90 & \underline{5.88} & 5.91 \\
 \bottomrule
\end{tabular}
\vspace{-3mm}
\end{table}

\begin{figure}[t]
  \centering
  \begin{subfigure}[t]{\columnwidth}
    \includegraphics[width=\columnwidth]{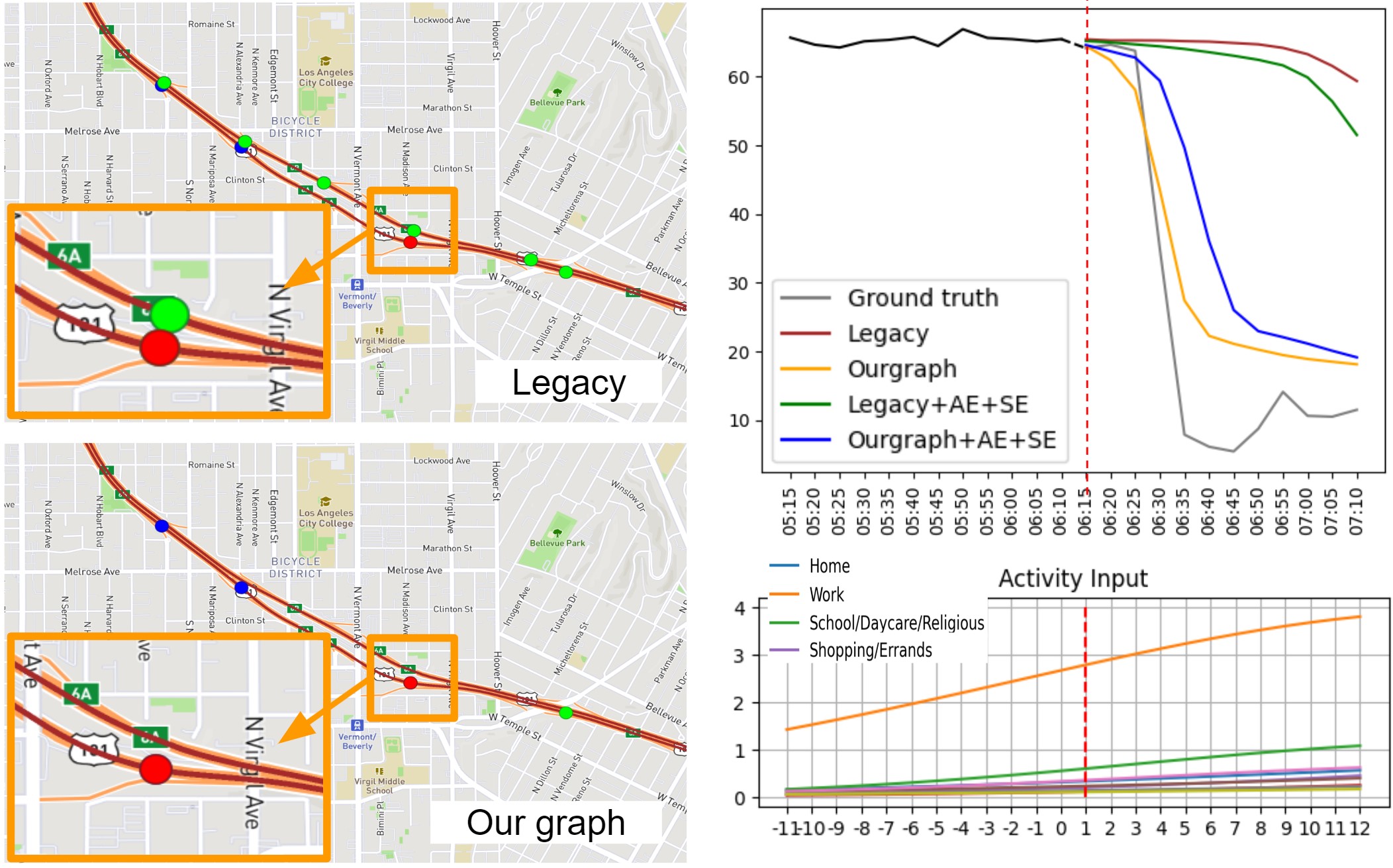}
    \caption{METR-LA, outperforms with our graph (ID: 716339)}
    \label{fig:case1}
  \end{subfigure}

  \vspace{\baselineskip}

  \begin{subfigure}[t]{\columnwidth}
    \includegraphics[width=\columnwidth]{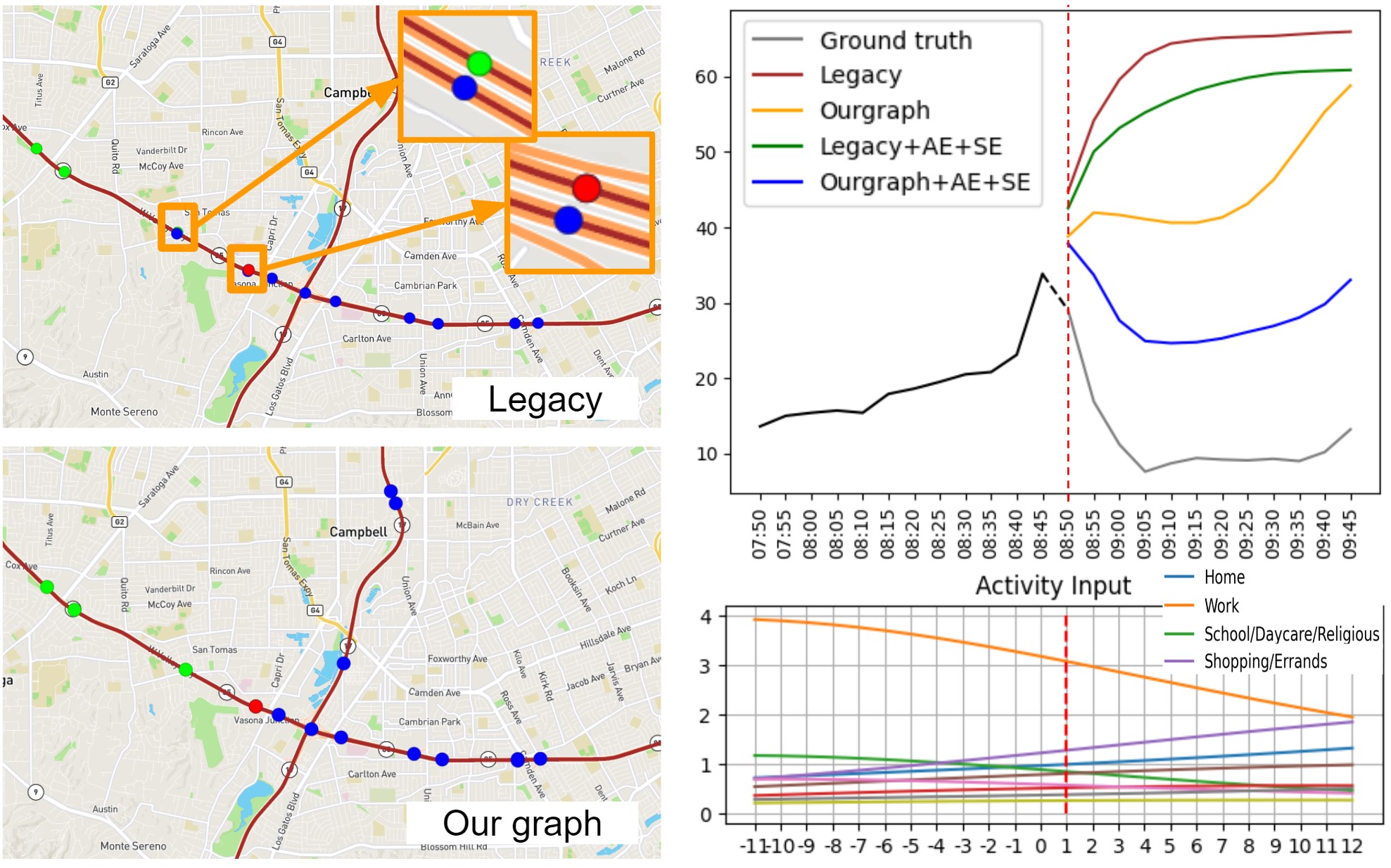}
    \caption{PEMS-BAY, outperforms with our graph with AE (ID: 400688)}
    \label{fig:case2}
  \end{subfigure}

\vspace{-2mm}

  \caption{Case study: UAGCRN on METR-LA and PEMS-BAY. Activity labels follows Fig.~\ref{fig:activity-embedding}. The target sensor (red), forward connected sensors (green), backward connected sensors (blue) are represented with colored markers on the map. Erroneous connections are found in legacy graphs.}
  \label{fig:comparison}
  
\vspace{-5mm}

\end{figure}

\subsection{Case Study}

Fig.\ref{fig:case1} and Fig.\ref{fig:case2} present a case study illustrating the superior performance of UA-GCRN$\dagger$ with our graph, sensor and activity embeddings. In both cases, the legacy graph includes incorrect connections that cannot be reached from the target sensor, which cause in wrong prediction.

In the METR-LA dataset (Fig.~\ref{fig:case1}), UA-GCRN$\dagger$ achieves better congestion prediction even without sensor and activity embeddings by accurately establishing connections between roads. This highlights the effectiveness of our approach in constructing the graph, which significantly improves the model's performance.

Furthermore, in the PEMS-BAY dataset (Fig.~\ref{fig:case2}), we observe that UA-GCRN$\dagger$ performs even better when provided with activity input. In this case, a high frequency of work activity is included in the historical sequence, and possible shopping activity in future sequence, which helps the model there can be consequent congestion in the prediction steps. The results demonstrate the additional benefit of incorporating activity information into the model, further enhancing its performance.

\subsection{Analysis of Sensor Reactions Based on Activity Information}

\begin{figure}[t]
  \includegraphics[width=\columnwidth]{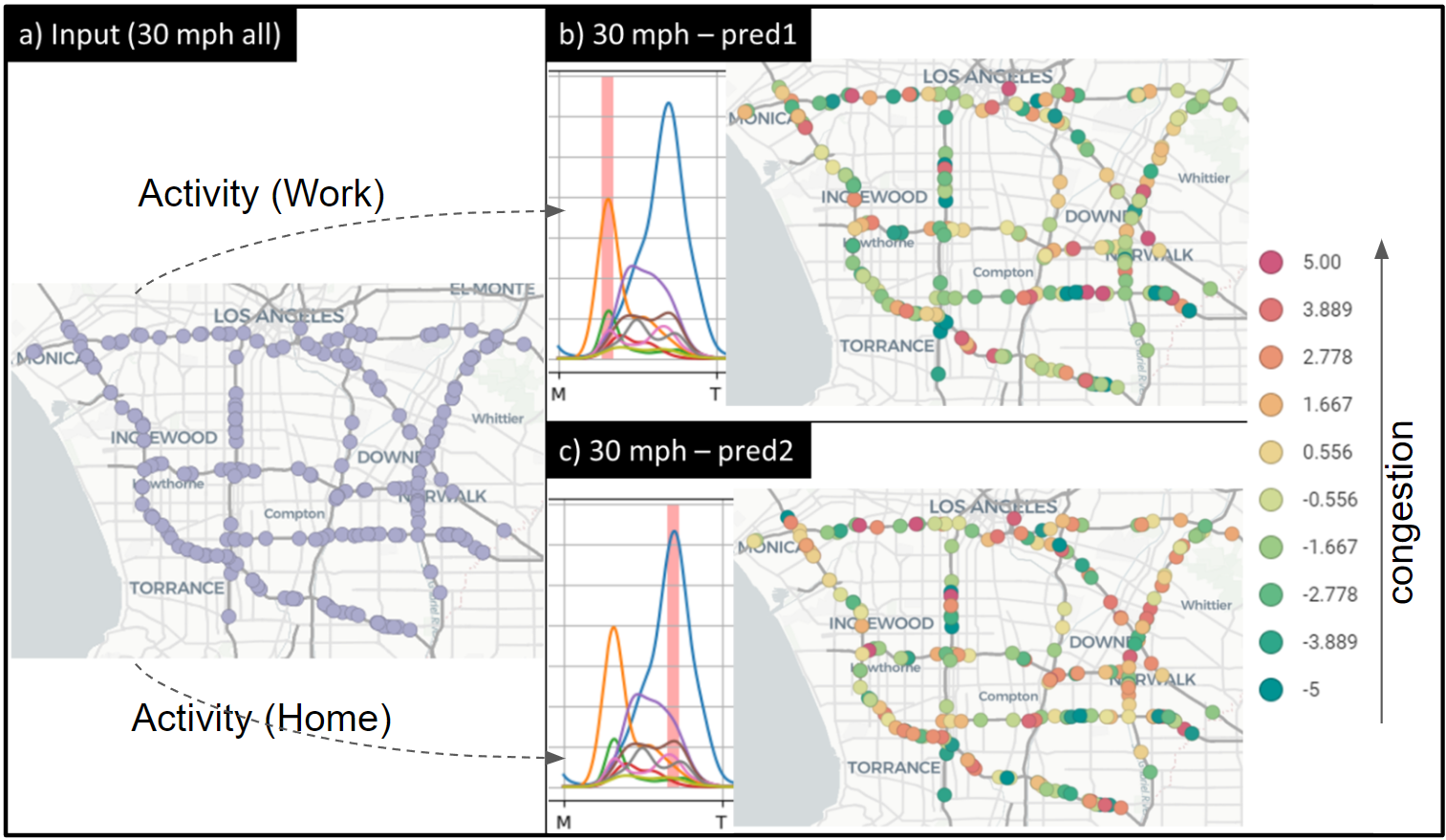}
  \vspace{-5mm}
  \caption{Sensor Reactions Based on Activity Information with UAGCRN (Red/Green: more/less congestion)}
  \label{fig:possible-simulation}
  \vspace{-5mm}
\end{figure}

We examine how sensors react differently when provided with different activity information. We conducted tests by setting all sensors to a speed value of 30 mph during the P sequence while varying the activity input, as illustrated in Fig.~\ref{fig:possible-simulation}. The choice of 30 mph is for testing whether congestion would increase or alleviate when the road capacity is full.
We conducted two predictions of next 15 min, pred1 and pred2, by providing activity information for the morning rush hour (6:35 to 8:20) and the evening commuting time (16:45 to 18:30), respectively.

Our findings revealed that sensors exhibited different behaviors based on the given activities. This discrepancy is due to varying levels of road utilization associated with specific activities. Notably, even when the same traffic values are given to the model, our model predicted distinct patterns as it had learned the sensor's typical response patterns corresponding to future activities.
Overall, these analyses highlight the importance of incorporating sensor embedding while inserting activity information into traffic prediction models, as it leads to a better understanding of sensor reactions and enhances the accuracy of congestion predictions influenced by urban human activity.

\section{Discussion}


Our current research focuses on enhancing activity-based traffic prediction models that consider spatial and temporal factors while incorporating travel purposes. However, there are notable areas for refinement. To enhance the realism of travel paths, it is essential to account for travel demands over time by integrating real-time data such as transportation and social media, enabling accurate inference of road-specific travel demands. Additionally, insights from land use, points of interest, and demographics can deepen our understanding of travel purposes and traffic patterns. Moreover, employing traffic simulation for synthetic data generation offers the potential to uncover nuanced traffic behaviors beyond our current A* algorithm-based approach. Exploring this dynamic route choice behavior using established transportation research and AI techniques could yield more effective predictions. By refining our methodology in these areas, future research can enhance the accuracy and applicability of our models, allowing us to better understand and predict traffic patterns in urban areas.


\section{Conclusion}

In conclusion, our research highlights the advantages of integrating real-world knowledge of urban human activity into spatiotemporal traffic prediction models. 
We propose a novel approach that effectively addresses the challenges of accurate graph construction, individual sensor heterogeneity handling, and human activity-based inference. 
Our proposed method incorporates realistic travel path generation with A* algorithm, co-occurrence and distance-based sensor connectivity measures, sensor-specific one-hot encoding, and human activity embedding into graph-convolution based spatiotemporal deep learning architectures. Experimental results demonstrate the effectiveness of our approach, surpassing other baselines and achieving state-of-the-art performance on real-world datasets. The insights gained from this study contribute to a better understanding of traffic patterns influenced by urban human activity, opening up avenues for further advancements in traffic prediction.


\begin{acks}
This work was supported by the Institute of Information \& Communications Technology Planning \& Evaluation (IITP) grant funded by the Korean government (MSIT) (No.2019-0-01126, Self-learning based Autonomic IoT Edge Computing).
\end{acks}

\newpage
\bibliographystyle{ACM-Reference-Format}
\bibliography{
    bib/uagcrn.bib
}


\begin{thebibliography}{33}


\ifx \showCODEN    \undefined \def \showCODEN     #1{\unskip}     \fi
\ifx \showDOI      \undefined \def \showDOI       #1{#1}\fi
\ifx \showISBNx    \undefined \def \showISBNx     #1{\unskip}     \fi
\ifx \showISBNxiii \undefined \def \showISBNxiii  #1{\unskip}     \fi
\ifx \showISSN     \undefined \def \showISSN      #1{\unskip}     \fi
\ifx \showLCCN     \undefined \def \showLCCN      #1{\unskip}     \fi
\ifx \shownote     \undefined \def \shownote      #1{#1}          \fi
\ifx \showarticletitle \undefined \def \showarticletitle #1{#1}   \fi
\ifx \showURL      \undefined \def \showURL       {\relax}        \fi
\providecommand\bibfield[2]{#2}
\providecommand\bibinfo[2]{#2}
\providecommand\natexlab[1]{#1}
\providecommand\showeprint[2][]{arXiv:#2}

\bibitem[PEM({[n.\,d.]})]%
        {PEMSWebpage}
 \bibinfo{year}{[n.\,d.]}\natexlab{}.
\newblock \bibinfo{title}{{California Performance Measurement System (PeMS)}}.
\newblock \bibinfo{howpublished}{\url{https://pems.dot.ca.gov/}}.
\newblock
\newblock
\shownote{Accessed: June 3, 2023}.


\bibitem[Ope({[n.\,d.]})]%
        {OpenStreetMap}
 \bibinfo{year}{[n.\,d.]}\natexlab{}.
\newblock \bibinfo{title}{OpenStreetMap}.
\newblock \bibinfo{howpublished}{\url{https://www.openstreetmap.org}}.
\newblock
\newblock
\shownote{Accessed: June 3, 2023}.


\bibitem[Axhausen and G{\"a}rling(1992)]%
        {axhausen1992activity}
\bibfield{author}{\bibinfo{person}{Kay~W Axhausen} {and} \bibinfo{person}{Tommy
  G{\"a}rling}.} \bibinfo{year}{1992}\natexlab{}.
\newblock \showarticletitle{Activity-based approaches to travel analysis:
  conceptual frameworks, models, and research problems}.
\newblock \bibinfo{journal}{\emph{Transport reviews}} \bibinfo{volume}{12},
  \bibinfo{number}{4} (\bibinfo{year}{1992}), \bibinfo{pages}{323--341}.
\newblock


\bibitem[Ben-Akiva and Bowman(1998)]%
        {ben1998activity}
\bibfield{author}{\bibinfo{person}{Moshe~E Ben-Akiva} {and}
  \bibinfo{person}{John~L Bowman}.} \bibinfo{year}{1998}\natexlab{}.
\newblock \showarticletitle{Activity based travel demand model systems}.
\newblock \bibinfo{journal}{\emph{Equilibrium and advanced transportation
  modelling}} (\bibinfo{year}{1998}), \bibinfo{pages}{27--46}.
\newblock


\bibitem[Bhat et~al\mbox{.}(2004)]%
        {bhat2004comprehensive}
\bibfield{author}{\bibinfo{person}{Chandra~R Bhat}, \bibinfo{person}{Jessica~Y
  Guo}, \bibinfo{person}{Sivaramakrishnan Srinivasan}, {and}
  \bibinfo{person}{Aruna Sivakumar}.} \bibinfo{year}{2004}\natexlab{}.
\newblock \showarticletitle{Comprehensive econometric microsimulator for daily
  activity-travel patterns}.
\newblock \bibinfo{journal}{\emph{Transportation Research Record}}
  \bibinfo{volume}{1894}, \bibinfo{number}{1} (\bibinfo{year}{2004}),
  \bibinfo{pages}{57--66}.
\newblock


\bibitem[Bowman and Ben-Akiva(2001)]%
        {bowman2001activity}
\bibfield{author}{\bibinfo{person}{John~L Bowman} {and}
  \bibinfo{person}{Moshe~E Ben-Akiva}.} \bibinfo{year}{2001}\natexlab{}.
\newblock \showarticletitle{Activity-based disaggregate travel demand model
  system with activity schedules}.
\newblock \bibinfo{journal}{\emph{Transportation research part a: policy and
  practice}} \bibinfo{volume}{35}, \bibinfo{number}{1} (\bibinfo{year}{2001}),
  \bibinfo{pages}{1--28}.
\newblock


\bibitem[Cai et~al\mbox{.}(2020)]%
        {TrafficTransformer}
\bibfield{author}{\bibinfo{person}{Ling Cai}, \bibinfo{person}{Krzysztof
  Janowicz}, \bibinfo{person}{Gengchen Mai}, \bibinfo{person}{Bo Yan}, {and}
  \bibinfo{person}{Rui Zhu}.} \bibinfo{year}{2020}\natexlab{}.
\newblock \showarticletitle{Traffic transformer: Capturing the continuity and
  periodicity of time series for traffic forecasting}.
\newblock \bibinfo{journal}{\emph{Trans. {GIS}}} \bibinfo{volume}{24},
  \bibinfo{number}{3} (\bibinfo{year}{2020}), \bibinfo{pages}{736--755}.
\newblock
\urldef\tempurl%
\url{https://doi.org/10.1111/tgis.12644}
\showDOI{\tempurl}


\bibitem[Castiglione et~al\mbox{.}(2015)]%
        {castiglione2015activity}
\bibfield{author}{\bibinfo{person}{Joe Castiglione}, \bibinfo{person}{Mark
  Bradley}, {and} \bibinfo{person}{John Gliebe}.}
  \bibinfo{year}{2015}\natexlab{}.
\newblock \bibinfo{booktitle}{\emph{Activity-based travel demand models: a
  primer}}.
\newblock Number SHRP 2 Report S2-C46-RR-1.
\newblock


\bibitem[Daganzo(1994)]%
        {daganzo1994cell}
\bibfield{author}{\bibinfo{person}{Carlos~F Daganzo}.}
  \bibinfo{year}{1994}\natexlab{}.
\newblock \showarticletitle{The cell transmission model: A dynamic
  representation of highway traffic consistent with the hydrodynamic theory}.
\newblock \bibinfo{journal}{\emph{Transportation research part B:
  methodological}} \bibinfo{volume}{28}, \bibinfo{number}{4}
  (\bibinfo{year}{1994}), \bibinfo{pages}{269--287}.
\newblock


\bibitem[Grover and Leskovec(2016)]%
        {node2vec}
\bibfield{author}{\bibinfo{person}{Aditya Grover} {and} \bibinfo{person}{Jure
  Leskovec}.} \bibinfo{year}{2016}\natexlab{}.
\newblock \showarticletitle{node2vec: Scalable feature learning for networks}.
  In \bibinfo{booktitle}{\emph{Proceedings of the 22nd ACM SIGKDD international
  conference on Knowledge discovery and data mining}}.
  \bibinfo{pages}{855--864}.
\newblock


\bibitem[Guo et~al\mbox{.}(2019)]%
        {ASTGCN}
\bibfield{author}{\bibinfo{person}{Shengnan Guo}, \bibinfo{person}{Youfang
  Lin}, \bibinfo{person}{Ning Feng}, \bibinfo{person}{Chao Song}, {and}
  \bibinfo{person}{Huaiyu Wan}.} \bibinfo{year}{2019}\natexlab{}.
\newblock \showarticletitle{Attention based spatial-temporal graph
  convolutional networks for traffic flow forecasting}. In
  \bibinfo{booktitle}{\emph{Proceedings of the AAAI conference on artificial
  intelligence}}, Vol.~\bibinfo{volume}{33}. \bibinfo{pages}{922--929}.
\newblock


\bibitem[Guo et~al\mbox{.}(2021)]%
        {ASTGCNNTF}
\bibfield{author}{\bibinfo{person}{Shengnan Guo}, \bibinfo{person}{Youfang
  Lin}, \bibinfo{person}{Huaiyu Wan}, \bibinfo{person}{Xiucheng Li}, {and}
  \bibinfo{person}{Gao Cong}.} \bibinfo{year}{2021}\natexlab{}.
\newblock \showarticletitle{Learning dynamics and heterogeneity of
  spatial-temporal graph data for traffic forecasting}.
\newblock \bibinfo{journal}{\emph{IEEE Transactions on Knowledge and Data
  Engineering}} \bibinfo{volume}{34}, \bibinfo{number}{11}
  (\bibinfo{year}{2021}), \bibinfo{pages}{5415--5428}.
\newblock


\bibitem[Hart et~al\mbox{.}(1968)]%
        {hart1968formal}
\bibfield{author}{\bibinfo{person}{Peter~E Hart}, \bibinfo{person}{Nils~J
  Nilsson}, {and} \bibinfo{person}{Bertram Raphael}.}
  \bibinfo{year}{1968}\natexlab{}.
\newblock \showarticletitle{A formal basis for the heuristic determination of
  minimum cost paths}.
\newblock \bibinfo{journal}{\emph{IEEE transactions on Systems Science and
  Cybernetics}} \bibinfo{volume}{4}, \bibinfo{number}{2}
  (\bibinfo{year}{1968}), \bibinfo{pages}{100--107}.
\newblock


\bibitem[Hochreiter and Schmidhuber(1997)]%
        {LSTM}
\bibfield{author}{\bibinfo{person}{Sepp Hochreiter} {and}
  \bibinfo{person}{J{\"u}rgen Schmidhuber}.} \bibinfo{year}{1997}\natexlab{}.
\newblock \showarticletitle{Long short-term memory}.
\newblock \bibinfo{journal}{\emph{Neural computation}} \bibinfo{volume}{9},
  \bibinfo{number}{8} (\bibinfo{year}{1997}), \bibinfo{pages}{1735--1780}.
\newblock


\bibitem[Jiang et~al\mbox{.}(2023)]%
        {PDFormer}
\bibfield{author}{\bibinfo{person}{Jiawei Jiang}, \bibinfo{person}{Chengkai
  Han}, \bibinfo{person}{Wayne~Xin Zhao}, {and} \bibinfo{person}{Jingyuan
  Wang}.} \bibinfo{year}{2023}\natexlab{}.
\newblock \showarticletitle{PDFormer: Propagation Delay-Aware Dynamic
  Long-Range Transformer for Traffic Flow Prediction}. In
  \bibinfo{booktitle}{\emph{Thirty-Seventh {AAAI} Conference on Artificial
  Intelligence, {AAAI} 2023, Thirty-Fifth Conference on Innovative Applications
  of Artificial Intelligence, {IAAI} 2023, Thirteenth Symposium on Educational
  Advances in Artificial Intelligence, {EAAI} 2023, Washington, DC, USA,
  February 7-14, 2023}}. \bibinfo{pages}{4365--4373}.
\newblock


\bibitem[Joubert and De~Waal(2020)]%
        {joubert2020activity}
\bibfield{author}{\bibinfo{person}{Johan~W Joubert} {and} \bibinfo{person}{Alta
  De~Waal}.} \bibinfo{year}{2020}\natexlab{}.
\newblock \showarticletitle{Activity-based travel demand generation using
  Bayesian networks}.
\newblock \bibinfo{journal}{\emph{Transportation Research Part C: Emerging
  Technologies}}  \bibinfo{volume}{120} (\bibinfo{year}{2020}),
  \bibinfo{pages}{102804}.
\newblock


\bibitem[Kitamura(1988)]%
        {kitamura1988evaluation}
\bibfield{author}{\bibinfo{person}{Ryuichi Kitamura}.}
  \bibinfo{year}{1988}\natexlab{}.
\newblock \showarticletitle{An evaluation of activity-based travel analysis}.
\newblock \bibinfo{journal}{\emph{Transportation}}  \bibinfo{volume}{15}
  (\bibinfo{year}{1988}), \bibinfo{pages}{9--34}.
\newblock


\bibitem[Li et~al\mbox{.}(2018)]%
        {DCRNN}
\bibfield{author}{\bibinfo{person}{Yaguang Li}, \bibinfo{person}{Rose Yu},
  \bibinfo{person}{Cyrus Shahabi}, {and} \bibinfo{person}{Yan Liu}.}
  \bibinfo{year}{2018}\natexlab{}.
\newblock \showarticletitle{Diffusion Convolutional Recurrent Neural Network:
  Data-Driven Traffic Forecasting}. In \bibinfo{booktitle}{\emph{6th
  International Conference on Learning Representations, {ICLR} 2018, Vancouver,
  BC, Canada, April 30 - May 3, 2018, Conference Track Proceedings}}.
\newblock


\bibitem[McNally(2000)]%
        {mcnally2000four}
\bibfield{author}{\bibinfo{person}{Michael~G McNally}.}
  \bibinfo{year}{2000}\natexlab{}.
\newblock \showarticletitle{THE FOUR-STEP MODEL. IN: HANDBOOK OF TRANSPORT
  MODELLING}.
\newblock  (\bibinfo{year}{2000}).
\newblock


\bibitem[Neutens et~al\mbox{.}(2011)]%
        {neutens2011prism}
\bibfield{author}{\bibinfo{person}{Tijs Neutens}, \bibinfo{person}{Tim
  Schwanen}, {and} \bibinfo{person}{Frank Witlox}.}
  \bibinfo{year}{2011}\natexlab{}.
\newblock \showarticletitle{The prism of everyday life: Towards a new research
  agenda for time geography}.
\newblock \bibinfo{journal}{\emph{Transport reviews}} \bibinfo{volume}{31},
  \bibinfo{number}{1} (\bibinfo{year}{2011}), \bibinfo{pages}{25--47}.
\newblock


\bibitem[Pinjari and Bhat(2011)]%
        {pinjari2011activity}
\bibfield{author}{\bibinfo{person}{Abdul~Rawoof Pinjari} {and}
  \bibinfo{person}{Chandra~R Bhat}.} \bibinfo{year}{2011}\natexlab{}.
\newblock \showarticletitle{Activity-based travel demand analysis}.
\newblock In \bibinfo{booktitle}{\emph{A handbook of transport Economics}}.
  \bibinfo{publisher}{Edward Elgar Publishing}.
\newblock


\bibitem[Schneider et~al\mbox{.}(2013)]%
        {schneider2013unravelling}
\bibfield{author}{\bibinfo{person}{Christian~M Schneider},
  \bibinfo{person}{Vitaly Belik}, \bibinfo{person}{Thomas Couronn{\'e}},
  \bibinfo{person}{Zbigniew Smoreda}, {and} \bibinfo{person}{Marta~C
  Gonz{\'a}lez}.} \bibinfo{year}{2013}\natexlab{}.
\newblock \showarticletitle{Unravelling daily human mobility motifs}.
\newblock \bibinfo{journal}{\emph{Journal of The Royal Society Interface}}
  \bibinfo{volume}{10}, \bibinfo{number}{84} (\bibinfo{year}{2013}),
  \bibinfo{pages}{20130246}.
\newblock


\bibitem[Seo et~al\mbox{.}(2018)]%
        {GCRN}
\bibfield{author}{\bibinfo{person}{Youngjoo Seo}, \bibinfo{person}{Micha{\"e}l
  Defferrard}, \bibinfo{person}{Pierre Vandergheynst}, {and}
  \bibinfo{person}{Xavier Bresson}.} \bibinfo{year}{2018}\natexlab{}.
\newblock \showarticletitle{Structured sequence modeling with graph
  convolutional recurrent networks}. In \bibinfo{booktitle}{\emph{Neural
  Information Processing: 25th International Conference, ICONIP 2018, Siem
  Reap, Cambodia, December 13-16, 2018, Proceedings, Part I 25}}. Springer,
  \bibinfo{pages}{362--373}.
\newblock


\bibitem[Shang et~al\mbox{.}(2021)]%
        {GTS}
\bibfield{author}{\bibinfo{person}{Chao Shang}, \bibinfo{person}{Jie Chen},
  {and} \bibinfo{person}{Jinbo Bi}.} \bibinfo{year}{2021}\natexlab{}.
\newblock \showarticletitle{Discrete Graph Structure Learning for Forecasting
  Multiple Time Series}. In \bibinfo{booktitle}{\emph{9th International
  Conference on Learning Representations, {ICLR} 2021, Virtual Event, Austria,
  May 3-7, 2021}}. \bibinfo{publisher}{OpenReview.net}.
\newblock
\urldef\tempurl%
\url{https://openreview.net/forum?id=WEHSlH5mOk}
\showURL{%
\tempurl}


\bibitem[Shao et~al\mbox{.}(2022)]%
        {STEP}
\bibfield{author}{\bibinfo{person}{Zezhi Shao}, \bibinfo{person}{Zhao Zhang},
  \bibinfo{person}{Fei Wang}, {and} \bibinfo{person}{Yongjun Xu}.}
  \bibinfo{year}{2022}\natexlab{}.
\newblock \showarticletitle{Pre-training enhanced spatial-temporal graph neural
  network for multivariate time series forecasting}. In
  \bibinfo{booktitle}{\emph{Proceedings of the 28th ACM SIGKDD Conference on
  Knowledge Discovery and Data Mining}}. \bibinfo{pages}{1567--1577}.
\newblock


\bibitem[Spear(1996)]%
        {spear1996new}
\bibfield{author}{\bibinfo{person}{Bruce~D Spear}.}
  \bibinfo{year}{1996}\natexlab{}.
\newblock \showarticletitle{New approaches to transportation forecasting
  models: A synthesis of four research proposals}.
\newblock \bibinfo{journal}{\emph{Transportation}}  \bibinfo{volume}{23}
  (\bibinfo{year}{1996}), \bibinfo{pages}{215--240}.
\newblock


\bibitem[Strehl and Ghosh(2002)]%
        {cooccur}
\bibfield{author}{\bibinfo{person}{Alexander Strehl} {and}
  \bibinfo{person}{Joydeep Ghosh}.} \bibinfo{year}{2002}\natexlab{}.
\newblock \showarticletitle{Cluster ensembles---a knowledge reuse framework for
  combining multiple partitions}.
\newblock \bibinfo{journal}{\emph{Journal of machine learning research}}
  \bibinfo{volume}{3}, \bibinfo{number}{Dec} (\bibinfo{year}{2002}),
  \bibinfo{pages}{583--617}.
\newblock


\bibitem[U.S. Department~of Transportation(2017)]%
        {nths-2017}
\bibfield{author}{\bibinfo{person}{Federal Highway~Administration U.S.
  Department~of Transportation}.} \bibinfo{year}{2017}\natexlab{}.
\newblock \showarticletitle{2017 National Household Travel Survey}.
\newblock  (\bibinfo{year}{2017}).
\newblock
\urldef\tempurl%
\url{http://nhts.ornl.gov}
\showURL{%
\tempurl}


\bibitem[Vaswani et~al\mbox{.}(2017)]%
        {transformer}
\bibfield{author}{\bibinfo{person}{Ashish Vaswani}, \bibinfo{person}{Noam
  Shazeer}, \bibinfo{person}{Niki Parmar}, \bibinfo{person}{Jakob Uszkoreit},
  \bibinfo{person}{Llion Jones}, \bibinfo{person}{Aidan~N. Gomez},
  \bibinfo{person}{Lukasz Kaiser}, {and} \bibinfo{person}{Illia Polosukhin}.}
  \bibinfo{year}{2017}\natexlab{}.
\newblock \showarticletitle{Attention is All you Need}. In
  \bibinfo{booktitle}{\emph{Advances in Neural Information Processing Systems
  30: Annual Conference on Neural Information Processing Systems 2017, December
  4-9, 2017, Long Beach, CA, {USA}}}. \bibinfo{pages}{5998--6008}.
\newblock


\bibitem[Wu et~al\mbox{.}(2019)]%
        {GraphWavenet}
\bibfield{author}{\bibinfo{person}{Zonghan Wu}, \bibinfo{person}{Shirui Pan},
  \bibinfo{person}{Guodong Long}, \bibinfo{person}{Jing Jiang}, {and}
  \bibinfo{person}{Chengqi Zhang}.} \bibinfo{year}{2019}\natexlab{}.
\newblock \showarticletitle{Graph wavenet for deep spatial-temporal graph
  modeling}.
\newblock \bibinfo{journal}{\emph{Proceedings of the Twenty-Eighth
  International Joint Conference on Artificial Intelligence (IJCAI-19)}}
  (\bibinfo{year}{2019}).
\newblock


\bibitem[Yu et~al\mbox{.}(2018)]%
        {STGCN}
\bibfield{author}{\bibinfo{person}{Bing Yu}, \bibinfo{person}{Haoteng Yin},
  {and} \bibinfo{person}{Zhanxing Zhu}.} \bibinfo{year}{2018}\natexlab{}.
\newblock \showarticletitle{Spatio-Temporal Graph Convolutional Networks: {A}
  Deep Learning Framework for Traffic Forecasting}. In
  \bibinfo{booktitle}{\emph{Proceedings of the Twenty-Seventh International
  Joint Conference on Artificial Intelligence, {IJCAI} 2018, July 13-19, 2018,
  Stockholm, Sweden}}, \bibfield{editor}{\bibinfo{person}{J{\'{e}}r{\^{o}}me
  Lang}} (Ed.). \bibinfo{publisher}{ijcai.org}, \bibinfo{pages}{3634--3640}.
\newblock
\urldef\tempurl%
\url{https://doi.org/10.24963/ijcai.2018/505}
\showDOI{\tempurl}


\bibitem[Zhao et~al\mbox{.}(2019)]%
        {T-GCN}
\bibfield{author}{\bibinfo{person}{Ling Zhao}, \bibinfo{person}{Yujiao Song},
  \bibinfo{person}{Chao Zhang}, \bibinfo{person}{Yu Liu}, \bibinfo{person}{Pu
  Wang}, \bibinfo{person}{Tao Lin}, \bibinfo{person}{Min Deng}, {and}
  \bibinfo{person}{Haifeng Li}.} \bibinfo{year}{2019}\natexlab{}.
\newblock \showarticletitle{T-gcn: A temporal graph convolutional network for
  traffic prediction}.
\newblock \bibinfo{journal}{\emph{IEEE Transactions on Intelligent
  Transportation Systems}} \bibinfo{volume}{21}, \bibinfo{number}{9}
  (\bibinfo{year}{2019}), \bibinfo{pages}{3848--3858}.
\newblock


\bibitem[Zheng et~al\mbox{.}(2020)]%
        {GMAN}
\bibfield{author}{\bibinfo{person}{Chuanpan Zheng}, \bibinfo{person}{Xiaoliang
  Fan}, \bibinfo{person}{Cheng Wang}, {and} \bibinfo{person}{Jianzhong Qi}.}
  \bibinfo{year}{2020}\natexlab{}.
\newblock \showarticletitle{{GMAN:} {A} Graph Multi-Attention Network for
  Traffic Prediction}. In \bibinfo{booktitle}{\emph{The Thirty-Fourth {AAAI}
  Conference on Artificial Intelligence, {AAAI} 2020, The Thirty-Second
  Innovative Applications of Artificial Intelligence Conference, {IAAI} 2020,
  The Tenth {AAAI} Symposium on Educational Advances in Artificial
  Intelligence, {EAAI} 2020, New York, NY, USA, February 7-12, 2020}}.
  \bibinfo{publisher}{{AAAI} Press}, \bibinfo{pages}{1234--1241}.
\newblock


\end{thebibliography}


\end{CJK}
\end{document}